\documentclass[10pt]{article}
\usepackage[accepted]{tmlr}

\usepackage{amsmath,amsfonts,bm}

\def\eqref#1{equation~\ref{#1}}
\def\1{\bm{1}}

\DeclareMathAlphabet{\mathsfit}{\encodingdefault}{\sfdefault}{m}{sl}
\SetMathAlphabet{\mathsfit}{bold}{\encodingdefault}{\sfdefault}{bx}{n}

\usepackage{hyperref}
\usepackage{url}
\usepackage[most]{tcolorbox}
\usepackage{xcolor}
\usepackage{booktabs}
\usepackage{threeparttable}
\usepackage{pifont}
\usepackage{enumitem}
\usepackage{multirow}
\usepackage{array}
\newsavebox{\dnarrowbox}
\AtBeginDocument{\savebox{\dnarrowbox}{$\downarrow$}}
\newcommand{\worse}{{\color{red}$\downarrow$}}

\usepackage{tikz}
\usepackage{pgf}
\usetikzlibrary{arrows.meta, positioning, fit, backgrounds, shapes.geometric, calc}

\setcitestyle{numbers,square,citesep={,},sort&compress}

\newcommand{\cmark}{\ding{51}}
\newcommand{\xmark}{\ding{55}}

\definecolor{boxbg}{RGB}{248,249,250}
\definecolor{boxborder}{RGB}{180,180,180}
\definecolor{mutedblue}{RGB}{76,110,155}
\hypersetup{
    colorlinks=true,
    linkcolor=blue,
    citecolor=mutedblue,
    urlcolor=mutedblue
}

\usepackage[table]{xcolor}

\newcommand{\perfbest}[1]{\cellcolor{green!18}#1}
\newcommand{\perfgood}[1]{\cellcolor{green!10}#1}
\newcommand{\perfmid}[1]{\cellcolor{yellow!15}#1}
\newcommand{\perfpoor}[1]{\cellcolor{orange!18}#1}
\newcommand{\perfworst}[1]{\cellcolor{red!15}#1}
\definecolor{indblue}{RGB}{37,99,235}
\definecolor{oodred}{RGB}{220,38,38}
\definecolor{boxbg}{RGB}{248,249,250}
\definecolor{boxborder}{RGB}{180,180,180}
\definecolor{mutedblue}{RGB}{76,110,155}
\hypersetup{
    colorlinks=true,
    linkcolor=blue,
    citecolor=mutedblue,
    urlcolor=mutedblue
}
\begin{document}

% \usepackage[most]{tcolorbox}
% \usepackage{xcolor}
% \usepackage{hyperref}
% \definecolor{boxbg}{RGB}{248,249,250}
% \definecolor{boxborder}{RGB}{180,180,180}

\noindent

\begin{center}
    \vspace*{4mm}

    \hspace{2mm}
    \begin{minipage}{0.8\textwidth}
        \centering
        {\LARGE \bfseries \textsf{
    \raisebox{-0.15\height}{\includegraphics[height=1.7em]{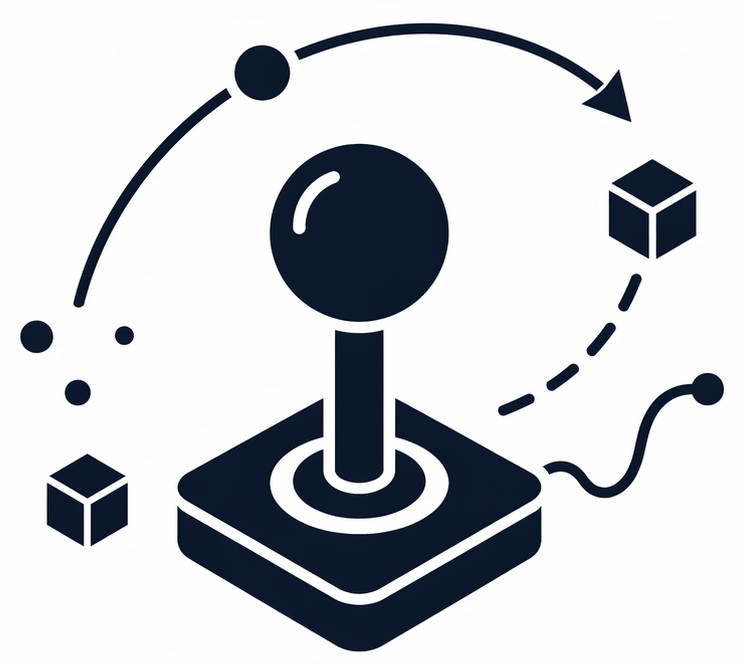}}
    ACWM-Phys:
}\par}
        {\Large \bfseries \textsf{Investigating Generalized Physical Interaction\\
        in Action-Conditioned Video World Models}\par}
    \end{minipage}

    \vspace{6mm}
    {\normalsize Haotian Xue$^{\dag}$, Yipu Chen$^*$, Liqian Ma$^*$, \\ Zelin Zhao, Lama Moukheiber, Yuchen Zhu, Yongxin Chen \par}
    \vspace{2mm}
    {\normalsize Georgia Institute of Technology\par}
    {$^{\dag}$ Project Lead; $^*$ equal contribution}
    \vspace{5mm}
    
    {
        \renewcommand{\arraystretch}{1.2}
        \begin{tabular}{@{} c @{\hskip 24pt} c @{\hskip 24pt} c @{}}
            \href{https://xavihart.github.io/ACWM-Phys}{%
                \raisebox{-3pt}{\includegraphics[height=16pt]{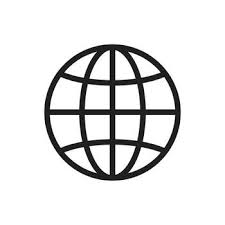}}%
                \hspace{4pt}\textsf{Project Page}%
            } &
            \href{https://huggingface.co/datasets/t1an/ACWM-Phys}{%
                \raisebox{-3pt}{\includegraphics[height=14pt]{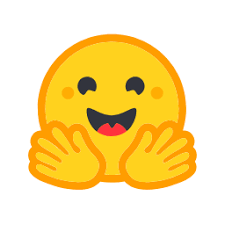}}%
                \hspace{4pt}\textsf{Hugging Face}%
            } &
            \href{https://github.com/xavihart/ACWM-Phys-dev}{%
                \raisebox{-3pt}{\includegraphics[height=14pt]{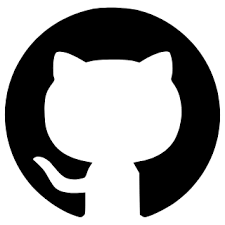}}%
                \hspace{4pt}\textsf{GitHub}%
            }
        \end{tabular}
    }
    \vspace{2mm}
\end{center}

\begin{abstract}
Action-conditioned world models (ACWMs) have shown strong promise for video prediction and decision-making. However, existing benchmarks are largely restricted to egocentric navigation or narrow, task-specific robotics datasets, offering only limited coverage of the rich physical interactions required for generalized world understanding. We introduce ACWM-Phys, a new benchmark for evaluating action-conditioned prediction under diverse physical dynamics in a clean, controllable simulation environment with a carefully designed action space. ACWM-Phys contains training and evaluation data spanning rigid-body dynamics, kinematics, deformable-object interactions, and particle dynamics. To evaluate both interpolation and generalization, we design in-distribution and out-of-distribution protocols with controlled shifts in interaction patterns or scene configurations. By building the benchmark in  a fully controllable simulator, ACWM-Phys enables precise data collection, reproducible evaluation, and systematic analysis of model capabilities for physically grounded world modeling. Through systematic experiments on ACWM-DiT, we find that OoD generalization depends not only on the physical regime but also on effective task complexity: models generalize well on visually simple, low-dimensional interactions with clear geometric structure, but suffer larger drops on deformable contacts, high-dimensional control, and complex articulated motion.  This suggests that the model still relies heavily on visual appearance patterns instead of fully learning the underlying physics. Ablations show that cross-attention improves high-dimensional action conditioning, causal VAEs outperform frame-wise encoders, and larger action spaces are harder to model but can improve generalization by providing richer control signals. These findings guide the design of physically grounded world models.
\end{abstract}
\vspace{-0.2cm}
\section{Introduction}

\begin{figure*}[t]
  \centering
  \includegraphics[width=\linewidth]{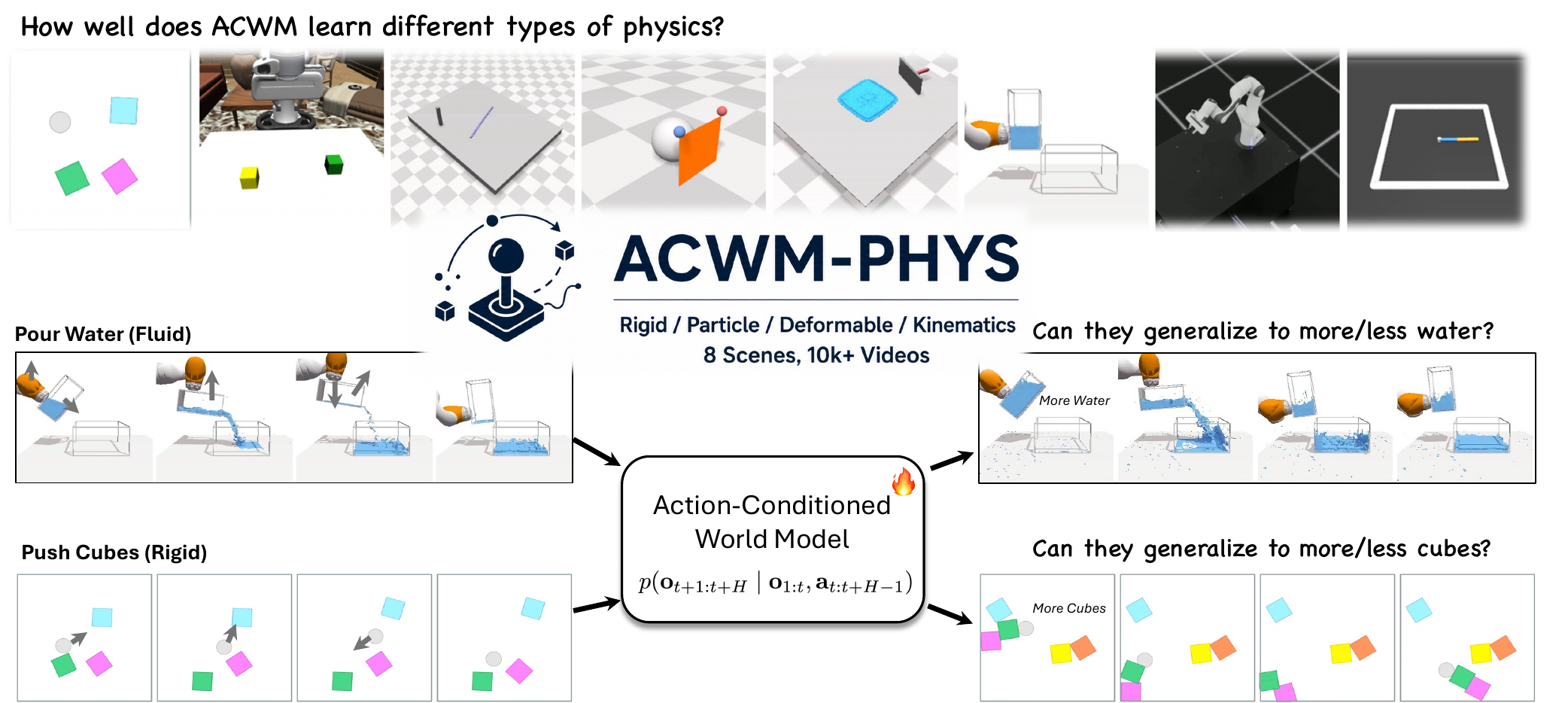}
  \caption{ACWM-Phys provides diverse physical scenes to help answer two questions: \textbf{how well can ACWMs learn different types of physics, and can they generalize beyond the training distribution?}
We evaluate both in-distribution prediction and out-of-distribution generalization, such as more/fewer water particles or cubes.}
  \label{fig:teaser}
  \vspace{-0.2cm}
\end{figure*}

Action-conditioned world models (ACWMs) have emerged as a promising paradigm for learning predictive models of the physical world from raw visual observations and control signals~\citep{dreamer4, huang2025vid2world, jiang2026wovr, parker2025genie, guo2025ctrl}.
By directly forecasting future observation sequences conditioned on agent actions, ACWMs hold the potential to serve as general-purpose simulators for robot planning, policy learning, and data augmentation without requiring hand-crafted dynamics models.
Recent advances in diffusion-based video generation~\cite{ho2020ddpm, karras2022edm, wan2025wan} have substantially improved visual fidelity and temporal consistency of generated sequences, further fueling interest in pixel-space world models that operate directly on high-dimensional observations.

Despite this progress, existing ACWMs and their accompanying benchmarks suffer from a critical blind spot: \emph{physical diversity}.
The vast majority of current work (Appendix Table~\ref{tab:benchmark_comparison}) is either confined to egocentric navigation~\cite{parker2025genie, RelicWorldModel2025, sun2025worldplay}, where actions correspond primarily to camera motion and objects rarely deform or interact, or to narrow robot manipulation~\cite{huang2025vid2world, chen2026bridgev2w, jiang2026wovr, guo2025ctrl} involving mostly rigid-body pick-and-place.
Yet the physical world encompasses a far richer spectrum of interaction regimes: deformable objects such as ropes and cloth, granular and fluid particle systems, and complex kinematic chains, each governed by fundamentally different dynamics.
It remains unclear whether current diffusion-based ACWMs can learn and generalize across these diverse interaction modes, or whether they silently fail when the underlying physics departs from the training distribution.

We address this gap with \textbf{ACWM-Phys}, a new benchmark designed to systematically evaluate ACWMs across four categories of physical interaction: rigid-body dynamics, deformable-object dynamics, particle dynamics, and kinematics.
ACWM-Phys comprises eight robotic simulation environments, each with carefully curated in-distribution (InD) training and test splits and a physically motivated out-of-distribution (OoD) test split that targets the specific generalization challenge most relevant to that environment such as unseen cloth sizes, doubled particle counts, or workspace regions excluded during training.
Because every environment is fully simulated, distribution shifts are exactly reproducible and free from sensor noise, enabling clean measurement of the generalization gap.

Alongside the benchmark, we introduce \textbf{ACWM-DiT}, a latent video diffusion transformer baseline.
ACWM-DiT builds on a pretrained causal video VAE~\cite{wan2025wan} for compact spatiotemporal encoding and couples a bidirectional DiT backbone~\cite{peebles2023scalable} with a  action-embedding module that inject action condition signal into pixel rendering.

Through systematic experiments on ACWM-Phys, we make the following contributions and findings:
\begin{itemize}[leftmargin=1.4em, itemsep=1pt]
  \item We introduce ACWM-Phys, the first benchmark spanning four distinct physical interaction regimes with controlled InD/OoD evaluation protocols across eight environments.
    \item We design ACWM-DiT as a strong diffusion-based baseline, which achieves strong performance across all environments and establishes a solid starting point for future work.
  \item We find that OoD generalization is driven primarily by task complexity rather than physics category: environments with low-dimensional geometric constraints (e.g., Push Cube, Reacher) generalize well, while tasks with high-DoF kinematics (Robot Arm) and contact-rich deformation (Cloth Move), indicating that models capture \textit{visual statistics} rather than \textit{physical laws}.
  \item Our ablations provide several design insights:(i) cross-attention conditioning outperforms AdaLN for high-dimensional action spaces but offers no benefit for low-dimensional actions; (ii) a causal video VAE with $4\times$ temporal compression outperforms a frame-independent encoder; and (iii) increasing the action-space dimensionality poses a greater learning challenge for the model, but it can also provide richer observational cues and thereby improve generalization for certain scenes.
\end{itemize}

\section{Related Works}
\paragraph{Action-conditioned World Models}
The idea of learning a model of the environment~\citep{2018worldmodel} for planning and decision-making has a long history in reinforcement learning. Recently, driven by rapid advances in diffusion-based image and video generation~\cite{ho2020ddpm, karras2022edm, wan2025wan, wan21github, huang2025selfforcing, yang2024cogvideox}, pixel-space world models have regained significant attention for generating high-quality visual predictions conditioned on actions~\citep{dreamer4, huang2025vid2world, RelicWorldModel2025, parker2025genie, ye2026world}. However, most existing works focus on egocentric settings, where actions primarily correspond to navigation, such as Genie-3~\cite{parker2025genie}, RELIC~\cite{RelicWorldModel2025}, and WorldPlay~\cite{sun2025worldplay}. These settings involve limited direct interaction with the environment. Other works instead concentrate on narrow domains, such as robot manipulation, including Vid2World~\cite{huang2025vid2world}, BridgeV2W~\cite{chen2026bridgev2w}, WoVR~\cite{jiang2026wovr}, and Ctrl-World~\cite{guo2025ctrl}, or on Minecraft gameplay~\cite{savva2026solaris, dreamer4}. A key limitation of these approaches is their limited investigation of complex physical interactions, as most mainly focus on simple navigation, or rigid-body dynamics such as picking, pushing, and grasping.

\paragraph{Physics in  Video Diffusion Models} Recent work has begun to investigate how well video diffusion models capture physical principles and whether they can serve as implicit world models~\cite{kang2024far, wang2025videoverse, motamed2026generative, zhang2025morpheus}, and further align current video diffusion to certain physics scenes~\cite{wang2025prophy, zhang2025thinkdiffusellmsguidedphysicsaware, yuan2026newtongen, le2025gravity}. These studies examine aspects such as physical law consistency~\cite{yuan2026newtongen, le2025gravity}, intuitive physics~\cite{wang2025prophy, li2025pisa}, and physical reasoning ability~\cite{zhang2025thinkdiffusellmsguidedphysicsaware, physinone} in generated videos, providing useful evidence on the current strengths and limitations of video generation models. However, most of this line of work remains centered on text-to-video (T2V) or image-to-video generation (T2V), where the model is asked to produce visually plausible dynamics from passive prompts or observations. As a result, these benchmarks primarily evaluate whether models can reflect physics in generated videos, rather than whether they can predict physically grounded futures under explicit action control. In contrast, our work focus on action-conditioned settings of ACWM.

\section{Background}
\label{sec:background}

\paragraph{Video Diffusion Models}
Video diffusion models~\cite{ho2022video, yang2024cogvideox, wan2025wan, wan21github} generate videos by transforming noise into data, typically conditioned on text, images, or other context. Given a video $\mathbf{x}\in\mathbb{R}^{T\times C\times H\times W}$ and condition $\mathbf{c}$, modern flow-matching formulations~\cite{liu2022flow, lipman2022flow} learn a time-dependent vector field $\mathbf{v}_\theta$ that transports noise $\mathbf{x}_0\sim p_0$ to data $\mathbf{x}_1\sim p_{\mathrm{data}}(\mathbf{x}\mid\mathbf{c})$ along a predefined path. In practice, this process is performed in a compressed latent space: a video encoder $\mathcal{E}$ maps $\mathbf{x}$ to $\mathbf{z}=\mathcal{E}(\mathbf{x})$, and a decoder $\mathcal{D}$ reconstructs $\mathbf{x}=\mathcal{D}(\mathbf{z})$. This reduces spatial-temporal cost and enables scalable Transformer-based denoisers such as DiT~\cite{peebles2023scalable}.

\paragraph{Action-Conditioned Video World Models}
Action-conditioned video world models extend video generation to controlled dynamics prediction. Given past observations $\mathbf{o}_{1:t}$ and future actions $\mathbf{a}_{t:t+H-1}$, the goal is to model
\[
p(\mathbf{o}_{t+1:t+H}\mid \mathbf{o}_{1:t}, \mathbf{a}_{t:t+H-1}),
\]
or equivalently in latent space,
\[
p(\mathbf{z}_{t+1:t+H}\mid \mathbf{z}_{1:t}, \mathbf{a}_{t:t+H-1}),
\qquad \mathbf{z}_t=\mathcal{E}(\mathbf{o}_t).
\]
Following recent diffusion-based ACWMs~\cite{huang2025vid2world, bagchi2026walk, jiang2026wovr}, we train the model to generate future latent trajectories conditioned on observation history and actions. Let $\mathbf{z}^{\mathrm{fut}}$ be the future latent video and $\mathbf{h}_t=\{\mathbf{z}_{1:t},\mathbf{a}_{t:t+H-1}\}$ be the conditioning context. Under flow matching, we sample $\mathbf{z}_0\sim\mathcal{N}(\mathbf{0},\mathbf{I})$, set $\mathbf{z}_1=\mathbf{z}^{\mathrm{fut}}$, interpolate
\[
\mathbf{z}_\tau=\alpha(\tau)\mathbf{z}_0+\beta(\tau)\mathbf{z}_1,
\]
and optimize
\[
\mathcal{L}_{\mathrm{ACWM}}=
\mathbb{E}
\left\|
\mathbf{v}_\theta(\mathbf{z}_\tau,\tau,\mathbf{h}_t)
-\dot{\alpha}(\tau)\mathbf{z}_0
-\dot{\beta}(\tau)\mathbf{z}_1
\right\|_2^2 .
\]
In our settings, we only use current frame as condition. After training, future observations are generated by integrating the learned vector field from noise while conditioning on past observations and candidate actions. In this work, we focus on the end-to-end rollout setting, where the model predicts the next fixed $H$ frames at once, while we can also do autoregressive generation like prior work~\cite{jiang2026wovr, guo2025ctrl} by iteratively denoising and conditioning on clean frames (in Appendix Figure~\ref{fig:ar_generation}).

\section{Investigating ACWMs for Learning Generalized Physical Interactions}

%% ─────────────────────────────────────────────────────────────────────────────
\subsection{ACWM-Phys: A Benchmark Suite for Rich Physical Interactions}

ACWM-Phys comprises eight robotic simulation environments grouped into four categories
of physical interaction: rigid-body dynamics, deformable-object dynamics, particle dynamics,
and kinematics (Figure~\ref{fig:dataset_overview}). Each category contains two environments
covering different object types, control spaces, and interaction patterns, ranging from
low-dimensional pushing and reaching tasks to contact-rich deformable and particle-based
dynamics. Each environment provides separate in-distribution (InD) training and test splits,
as well as an out-of-distribution (OoD) test split with a controlled distribution shift
along a physically meaningful axis, such as object count, workspace range, rope/cloth size,
particle quantity, or goal region. In total, the benchmark contains more than 15k simulated
trajectories with paired image observations, actions, and evaluation labels. Dataset sizes,
action specifications, and split details are provided in Appendix~\ref{app:dataset_stats}.
\input{figures/dataset_overview_fig}

\subsubsection{Categories of Physical Interactions}

\textbf{Rigid-Body Dynamics.}
\texttt{Push Cube} moves one to five colored cubes using a circular pusher, where
$\mathbf{a}\in\mathbb{R}^{2}$ specifies the pusher's absolute 2D target position.
\texttt{Stack Cube} uses a Franka Panda to place a red cube on a green cube, with
$\mathbf{a}\in\mathbb{R}^{7}$ denoting delta 6-DoF end-effector pose plus gripper command.

\textbf{Deformable-Object Dynamics.}
\texttt{Push Rope} uses a pole pusher to deform a flexible rope in PyFlex~\cite{li2018learning},
with $\mathbf{a}\in\mathbb{R}^{2}$ as the pole's horizontal displacement.
\texttt{Cloth Move} pushes a cloth over a fixed sphere using dual arms, with a shared
3D end-effector displacement $\mathbf{a}\in\mathbb{R}^{3}$; we study the full 8-D
per-arm action space in the ablation.

\textbf{Particle Dynamics.}
\texttt{Push Sand} rearranges granular material in PyFleX using a board pusher, with
$\mathbf{a}\in\mathbb{R}^{7}$ encoding the board's 3D pose delta.
\texttt{Pour Water} pours fluid by moving and tilting a cup, where
$\mathbf{a}\in\mathbb{R}^{4}$ gives Cartesian and tilt-angle deltas from a spring-damper controller.

\textbf{Kinematics.}
\texttt{Robot Arm} uses a 7-DoF Franka Panda in Isaac Sim with cuRobo planning, where
$\mathbf{a}\in\mathbb{R}^{7}$ is the per-joint angle delta.
\texttt{Reacher} controls a two-link MuJoCo~\cite{todorov2012mujoco} arm, with
$\mathbf{a}\in\mathbb{R}^{2}$ directly specifying joint torques. Please refer to Appendix Figure~\ref{fig:dataset_viz} for visualizations of scene rollouts.

\subsubsection{In-Distribution and Out-of-Distribution Evaluation Protocols}

A central design principle of ACWM-Phys is that \emph{every} environment supports
a controlled, physically motivated distribution shift between the InD and OoD splits.
Rather than applying random perturbations, we shift the physical parameters or
workspace regions that most directly challenge the generalization of a learned world model, detailed design of OoD scenes are in Appendx~\ref{app:ood-design}:

\begin{itemize}[leftmargin=1.4em, itemsep=1pt]
  \item \textbf{Rigid}: Push Cube tests unseen cube counts; Stack Cube shifts target placement.
  \item \textbf{Deformable}: Push Rope changes rope length; Cloth Move varies cloth size.
  \item \textbf{Particle}: Push Sand increases particle count; Pour Water shifts water level.
  \item \textbf{Kinematics}: Robot Arm expands the goal workspace; Reacher tests unseen goal regions.
\end{itemize}

Because all environments are fully simulated, OoD shifts are exactly reproducible and free from sensor noise, enabling precise measurement of generalization gaps, unlike real-robot benchmarks~\cite{khazatsky2024droid}. Models are trained only on InD data and evaluated on both InD and OoD test trajectories. We report MSE, SSIM~\cite{psnr}, and PSNR, and additionally use Masked-MSE (M-MSE), which computes MSE only on pixels with sufficient ground-truth temporal change, emphasizing motion-relevant regions while down-weighting static backgrounds; see Appendix~\ref{app:mmse}.
%% ─────────────────────────────────────────────────────────────────────────────

\begin{figure}[t]
  \centering
  \begin{minipage}[c]{0.53\linewidth}
    \centering
    \includegraphics[width=\linewidth]{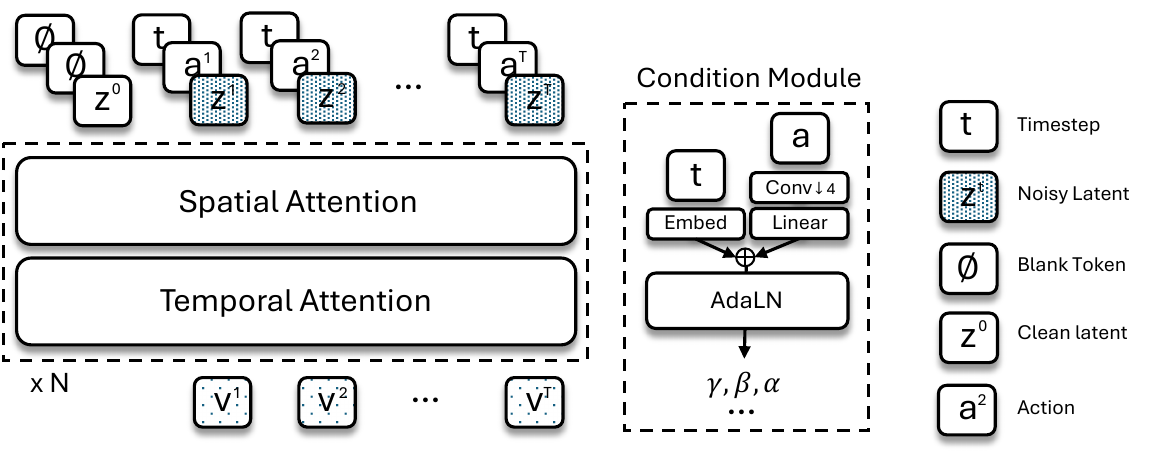}
  \end{minipage}
  \hspace{0.02\linewidth}
  \begin{minipage}[c]{0.40\linewidth}
    \caption{%
      \textbf{ACWM-DiT architecture.}
      Noisy latent tokens $\mathbf{z}_{1:T_l}$ (conditioning frames at $\sigma{=}0$,
      predicted frames at diffusion step $\sigma$) are processed by $N$ stacked DiT blocks
      with alternating spatial and temporal self-attention, modulated via AdaLN from a
      joint conditioning signal formed by
      summing the timestep embedding and the temporally compressed action embedding.
    }
    \label{fig:model_arch}
  \end{minipage}
  \vspace{-0.5cm}
\end{figure}

\subsection{ACWM-DiT: An Action-Conditioned Video Diffusion Transformer in Latent Space}
\label{sec:acwm-dit}

We use a DiT-based latent video diffusion model as a reproducible baseline for action-conditioned world modeling, named ACWM-DiT. Our goal is not to propose a new architecture, but to provide a strong and standardized diffusion-based baseline for diagnosing physical generalization on ACWM-Phys. As shown in Figure~\ref{fig:model_arch}, ACWM-DiT encodes video observations with a frozen WanVAE~\cite{wan2025wan} and denoises future latent tokens using a bidirectional DiT backbone with interleaved spatial and temporal self-attention and RoPE positional encoding. Actions are embedded by an MLP followed by a strided temporal convolution, which downsamples pixel-rate actions to the latent temporal resolution. The resulting action embeddings are injected into every transformer block through AdaLN as a joint action--timestep conditioning signal. The first frame is always kept clean and used as the history input, while the model predicts the remaining future frames in latent space. More architectural and training details are provided in Appendix~\ref{sec:app-model-structure}.
\vspace{-0.2cm}

\section{Experiments}
\label{sec:experiments}

%% ─────────────────────────────────────────────────────────────────────────────
\paragraph{Training Setup} All models are trained seperately from scratch with the AdamW optimizer at a learning rate of $10^{-4}$
with gradient clipping at 1.0.
We train for 100k steps with a batch size of 4 on 8 H100 GPUs for each task.
The flow-matching scheduler uses 1000 noise levels with a shift parameter $s{=}5.0$
and a Gaussian weighting envelope centered at noise step 500 to focus supervision
on intermediate denoising levels.
Video observations are encoded by the frozen Wan~2.1 causal VAE into latent tokens of
spatial resolution $H/8{\times}W/8$ with 16 channels and $4\times$ temporal compression.
Input sequences are padded/trimmed to a fixed latent length of $T_l{=}37$ tokens.
All environments are resized to $240{\times}240$ pixels prior to encoding,
except Push Sand which uses $240{\times}400$ to preserve its landscape aspect ratio.
All environments are trained with the AdaLN action-conditioning variant; cross-attention conditioning is evaluated separately in the ablation studies (Section \ref{sec:experiments}).

%% ─────────────────────────────────────────────────────────────────────────────
\subsection{Main Results}

Table~\ref{tab:eval_main} reports ACWM-DiT-S on all eight ACWM-Phys environments
at 100k training steps and 50 inference steps. We study the effect of sampling steps in Appendix~\ref{app:steps_curves}
.% General evaluation results: ACWM-DiT-S, steps=50, 100k training steps
\begin{table*}[t]
\centering
\label{tab:eval_main}
\small
\setlength{\tabcolsep}{4pt}
\resizebox{\textwidth}{!}{%
\begin{threeparttable}
\begin{tabular}{ll cccc cccc}
\toprule
& &
\multicolumn{4}{c}{\textbf{In-Distribution Test}} &
\multicolumn{4}{c}{\textbf{Out-of-Distribution Test}} \\
\cmidrule(lr){3-6}\cmidrule(lr){7-10}
\textbf{Category} & \textbf{Environment} &
  MSE$\downarrow$ & M-MSE$\downarrow$ & SSIM$\uparrow$ & PSNR$\uparrow$ &
  MSE$\downarrow$ & M-MSE$\downarrow$ & SSIM$\uparrow$ & PSNR$\uparrow$ \\
\midrule
\multirow{2}{*}{Rigid}
  & \perfmid{Push Cube}  & 2.919 & 11.43 & 0.955 & 25.35 & 2.950\worse  & \phantom{0}7.47        & 0.954\worse & 25.30\worse \\
  & \perfmid{Stack Cube} & 5.516 & 10.93 & 0.889 & 22.58 & 6.996\worse  & 13.40\worse & 0.872\worse & 21.55\worse \\
\midrule
\multirow{2}{*}{Deformable}
  & \perfbest{Push Rope} & 0.214  & \phantom{0}2.61 & 0.988 & 36.70 & 0.329\worse  & 12.86\worse & 0.985\worse & 34.83\worse \\
  & \perfworst{Cloth Move} & 10.667 & 63.68 & 0.920 & 19.72 & 23.820\worse & 93.67\worse & 0.864\worse & 16.23\worse \\
\midrule
\multirow{2}{*}{Particle}
  & \perfgood{Push Sand}  & 0.519 & \phantom{0}6.22 & 0.975 & 32.85 & 1.526\worse & 15.54\worse & 0.941\worse & 28.16\worse \\
  & \perfpoor{Pour Water} & 2.630 & 11.81 & 0.911 & 25.80 & 3.488\worse & 14.21\worse & 0.874\worse & 24.57\worse \\
\midrule
\multirow{2}{*}{Kinematics}
  & \perfpoor{Robot Arm} & 1.434 & 13.45 & 0.969 & 28.43 & 6.559\worse & 53.80\worse & 0.902\worse & 21.83\worse \\
  & \perfbest{Reacher}   & 0.260 & \phantom{0}5.63 & 0.992 & 35.85 & 0.272\worse & \phantom{0}8.89\worse & 0.992       & 35.65\worse \\
\bottomrule
\end{tabular}
\end{threeparttable}
}
\caption{%
  \textbf{ACWM-DiT-S evaluation on ACWM-Phys} (diffusion steps\,=\,50, 100k training steps).
  MSE and Masked-MSE (M-MSE) are scaled by $10^{-3}$.
   Environment names are shaded from green to red to qualitatively indicate overall prediction difficulty/performance, where greener backgrounds denote better model performance and redder backgrounds denote harder environments with larger prediction errors.
  {\color{red}$\downarrow$}~denotes OoD performance worse than InD.
}
\vspace{-0.4cm}
\end{table*}
\vspace{-0.2cm}

\paragraph{InD performance.}
ACWM-DiT-S achieves strong in-distribution performance across all four physics categories. Environments with simpler, repetitive dynamics achieve the highest fidelity: Push Rope (M-MSE 2.61, SSIM 0.988) and Reacher (M-MSE 5.63, SSIM 0.992) are predicted with near-perfect structural similarity and low motion-region error, indicating that the model captures both dynamic foreground behavior and global spatial coherence. Stack Cube (M-MSE 10.93, SSIM 0.889) and Cloth Move (M-MSE 63.68, SSIM 0.920) pose the greatest challenge: the lower InD SSIM for Stack Cube is mainly due to large foreground motion of the robot arm, which introduces substantial dynamic changes across frames, while Cloth Move's substantially higher M-MSE shows that large-scale deformation leads to much larger errors in physically dynamic regions even within the training distribution. 
\vspace{-0.2cm}

\paragraph{OoD generalization.}
Under distribution shift, ACWM-DiT-S shows consistent degradation, especially in motion-sensitive regions. The largest drops occur for Robot Arm ($\Delta$M-MSE $=+40.35$, $\Delta$SSIM $=-0.067$) and Cloth Move ($\Delta$M-MSE $=+29.99$, $\Delta$SSIM $=-0.056$), where unseen articulated configurations and large-scale cloth deformation introduce complex motion beyond the training distribution. This suggests that the model still relies partly on learned visual regularities rather than fully internalizing general physical dynamics.

Overall, OoD robustness is shaped by both physical complexity and action/state dimensionality. Push Cube ($\Delta$SSIM $=-0.001$) and Reacher ($\Delta$SSIM $=0.000$) remain nearly stable because their dynamics follow low-dimensional geometric constraints. Push Sand shows increased motion-region error ($\Delta$M-MSE $=+9.32$) while retaining moderate structural similarity (OoD SSIM $=0.941$, $\Delta$SSIM $=-0.034$), indicating difficulty in fine-grained particle redistribution. Pour Water is more stable in M-MSE ($\Delta$M-MSE $=+2.40$), likely because the pouring trajectory is repeatable, although SSIM still drops under unseen water volumes ($\Delta$SSIM $=-0.037$).

\begin{figure}[h]
  \centering
  \definecolor{indblue}{RGB}{37,99,235}
\definecolor{oodred}{RGB}{220,38,38}

\newcommand{\imgbox}[2]{%
  \fbox{\includegraphics[width=#1]{#2}}%
}

{\setlength{\tabcolsep}{1pt}%
\renewcommand{\arraystretch}{0.85}%
\setlength{\fboxsep}{0pt}%
\setlength{\fboxrule}{0.4pt}%
\scriptsize%
\begin{tabular}{@{}r@{\hspace{3pt}}cc@{}}
  \multicolumn{3}{@{}l@{}}{\textcolor{indblue}{\textbf{In-Distribution}}} \\[1pt]
  \rotatebox{90}{\textbf{GT}}
    & \imgbox{0.47\linewidth}{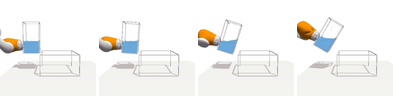}
    & \imgbox{0.47\linewidth}{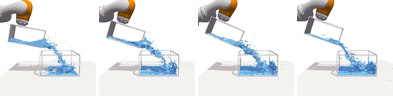} \\[0.5pt]
  \rotatebox{90}{\textbf{Pred}}
    & \imgbox{0.47\linewidth}{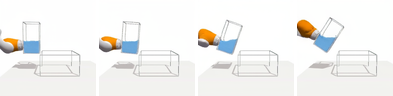}
    & \imgbox{0.47\linewidth}{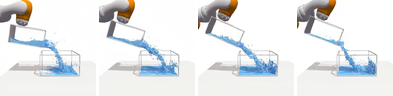} \\[4pt]
  \multicolumn{3}{@{}l@{}}{\textcolor{oodred}{\textbf{Out-of-Distribution}}} \\[1pt]
  \rotatebox{90}{\textbf{GT}}
    & \imgbox{0.47\linewidth}{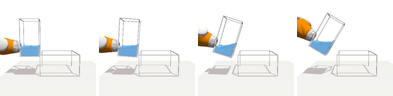}
    & \imgbox{0.47\linewidth}{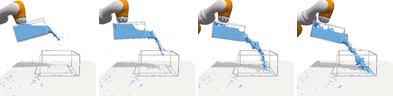} \\[0.5pt]
  \rotatebox{90}{\textbf{Pred}}
    & \imgbox{0.47\linewidth}{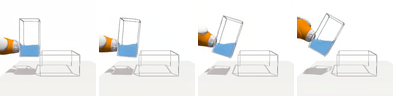}
    & \imgbox{0.47\linewidth}{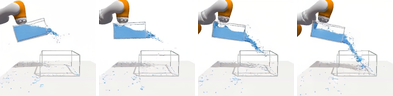} \\
\end{tabular}}
  \caption{%
    \textbf{Case study: Pour Water.}
    GT (top) and predicted (bottom) frames at four evenly-spaced timesteps.
    Two InD episodes (top block) and two OoD episodes (bottom block) with less water (left) and more water (right);
   The robot arm closely follows the ground-truth trajectory, indicating accurate prediction of articulated motion. Pour Water is also predicted well overall, although in the OoD setting the model sometimes underestimates the water amount, causing part of the fluid to disappear.
  }
\vspace{-0.2cm}
  
  \label{fig:case_pour_water}
\end{figure}

\begin{figure}[h]
  \centering

\newcommand{\imgbox}[2]{%
  \fbox{\includegraphics[width=#1]{#2}}%
}

{\setlength{\tabcolsep}{1pt}%
\renewcommand{\arraystretch}{0.85}%
\setlength{\fboxsep}{0pt}%
\setlength{\fboxrule}{0.4pt}%
\scriptsize%
\begin{tabular}{@{}r@{\hspace{3pt}}cc@{}}
  \multicolumn{3}{@{}l@{}}{\textcolor{indblue}{\textbf{In-Distribution}}} \\[1pt]
  \rotatebox{90}{\textbf{GT}}
    & \imgbox{0.47\linewidth}{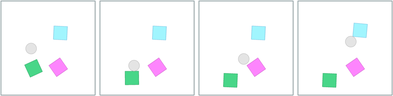}
    & \imgbox{0.47\linewidth}{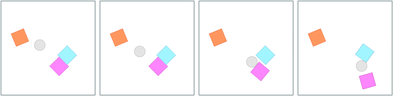} \\[0.5pt]
  \rotatebox{90}{\textbf{Pred}}
    & \imgbox{0.47\linewidth}{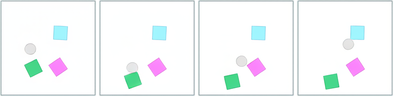}
    & \imgbox{0.47\linewidth}{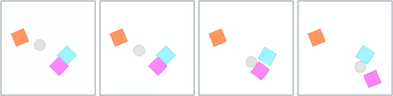} \\[4pt]
  \multicolumn{3}{@{}l@{}}{\textcolor{oodred}{\textbf{Out-of-Distribution}}} \\[1pt]
  \rotatebox{90}{\textbf{GT}}
    & \imgbox{0.47\linewidth}{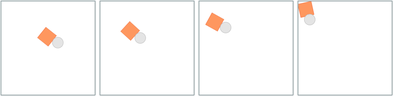}
    & \imgbox{0.47\linewidth}{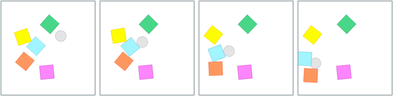} \\[0.5pt]
  \rotatebox{90}{\textbf{Pred}}
    & \imgbox{0.47\linewidth}{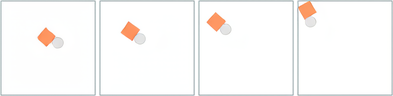}
    & \imgbox{0.47\linewidth}{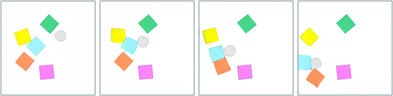} \\
\end{tabular}}
  \caption{%
    \textbf{Case study: Push Cube.}
    GT (top) and predicted (bottom) frames at four evenly-spaced timesteps.
    Two InD episodes (top block) and two OoD episodes (bottom block) show
    diverse cube configurations, with one cube  (left) and four cubes (right).
    The model accurately tracks cube positions and push trajectories
    across both distributions.
  }
  \label{fig:case_push_cube}
\vspace{-0.5cm}
  
\end{figure}
\paragraph{Case study: Pour Water.}

The model correctly predicts the pouring trajectory and the general fluid dynamics
for in-distribution water levels (Figure~\ref{fig:case_pour_water} in Appendix).
Under OoD shifts (fewer or far more water layers than seen during training), the model
tends to generate visually plausible but physically inperfect fill levels,
highlighting the gap between perceptual quality and true physical understanding.

\paragraph{Case study: Push Cubes.}
Figure~\ref{fig:case_push_cube} contrasts two OoD regimes: a single cube pushed to
an out-of-distribution workspace position, and a scene with 4+ cubes (unseen during training).
The model accurately tracks rigid-body trajectories in InD cases and generalizes well to OoD settings overall, although cubes occasionally disappear abruptly in some OoD videos, as shown in the appendix visualization.

More per-environment case studies for all remaining environments are provided
in Appendix~\ref{app:case_studies}.
\vspace{-0.2cm}

\paragraph{Generalization summary.}
Across environments, OoD generalization is shaped by both physical complexity and action/state dimensionality. Tasks with low-dimensional, visually clear geometric structure, such as rigid-body translation or simple joint trajectories, transfer more reliably to unseen configurations. In contrast, contact-rich deformation, particle dynamics, and high-DoF control lead to larger degradation, suggesting that current diffusion-based world models still rely heavily on appearance statistics rather than fully learning physical structure.
\vspace{-0.2cm}

%% ─────────────────────────────────────────────────────────────────────────────
\subsection{Ablation Studies}

\definecolor{lskyblue}{RGB}{135,206,250}
\newcommand{\finding}[1]{%
  \begin{tcolorbox}[
    colback=lskyblue!22!white,
    colframe=lskyblue!55!white,
    boxrule=0.5pt, arc=4pt,
    left=7pt, right=7pt, top=4pt, bottom=4pt,
    before skip=4pt, after skip=6pt,
  ]
  \small #1
  \end{tcolorbox}%
}

We conduct ablation studies along the following axes to provide further insights:
model scale, action-conditioning mechanism, latent-space formulation,
training data volume, and action dimensionality.

%% ── 1. Model Scale ────────────────────────────────────────────────────────────
\finding{\textbf{Model Scale.} Scaling model capacity consistently improves OoD robustness more than InD accuracy,
suggesting larger models better internalize physical structure rather than memorizing appearance statistics.
Gains diminish from DiT-M to DiT-L at the current data scale.}
Table~\ref{tab:abl_scale} compares
DiT-S (${\approx}200$M), DiT-M (${\approx}600$M), and DiT-L (${\approx}800$M)
on Cloth Move (deformable) and Robot Arm (kinematics),
two environments that represent qualitatively different physical interaction regimes.
Scaling from DiT-S to DiT-M consistently improves both InD and OoD performance,
with larger gains on OoD, suggesting that model capacity helps internalize physical
structure rather than merely memorizing training appearances.
Gains from DiT-M to DiT-L are more modest, indicating diminishing returns at this data scale.

\begin{table}[h]
\centering
\caption{%
  \textbf{Model scale ablation} on Cloth Move (3-DoF action, 50k steps) and Robot Arm (50k steps).
}
\label{tab:abl_scale}
\small
\setlength{\tabcolsep}{5pt}
\begin{tabular}{lc cccc cccc}
\toprule
& &
\multicolumn{4}{c}{\textbf{Cloth Move}} &
\multicolumn{4}{c}{\textbf{Robot Arm}} \\
\cmidrule(lr){3-6}\cmidrule(lr){7-10}
\textbf{Model} & \textbf{Params} &
  \multicolumn{2}{c}{InD} & \multicolumn{2}{c}{OoD} &
  \multicolumn{2}{c}{InD} & \multicolumn{2}{c}{OoD} \\
\cmidrule(lr){3-4}\cmidrule(lr){5-6}\cmidrule(lr){7-8}\cmidrule(lr){9-10}
& & MSE$\downarrow$ & PSNR$\uparrow$ & MSE$\downarrow$ & PSNR$\uparrow$
  & MSE$\downarrow$ & PSNR$\uparrow$ & MSE$\downarrow$ & PSNR$\uparrow$ \\
\midrule
DiT-S & ${\approx}200$M & 10.776 & 19.68 & 22.416 & 16.49 & 1.452 & 28.38 & 6.552 & 21.84 \\
DiT-M & ${\approx}600$M & 10.248 & 19.89 & 20.031 & 16.98 & 1.247 & 29.04 & 4.891 & 23.11 \\
DiT-L & ${\approx}800$M &  9.985 & 20.01 & 18.876 & 17.24 & 1.183 & 29.27 & 4.453 & 23.51 \\
\bottomrule
\end{tabular}
\end{table}

%% ── 2. Action Conditioning ────────────────────────────────────────────────────
\finding{\textbf{Action Conditioning.} Cross-attention is mainly beneficial for high-dimensional action spaces. For low-dimensional controls ($d_a{=}2$), AdaLN performs similarly or slightly better, while cross-attention gives clear gains on Robot Arm and modest OoD gains on Cloth Move.}

We compare AdaLN-based action conditioning with a cross-attention variant that injects action tokens through dedicated cross-attention layers. As shown in Table~\ref{tab:abl_action_cond}, cross-attention brings no benefit on Push Cube and Push Rope ($d_a{=}2$), where AdaLN already captures simple displacement controls effectively. In contrast, for Robot Arm ($d_a{=}7$), cross-attention substantially improves both InD and OoD performance, suggesting better binding between joint commands and articulated motion. For Cloth Move ($d_a{=}8$), cross-attention slightly hurts InD performance but modestly improves OoD performance. Overall, cross-attention is most useful when actions are high-dimensional and require structured spatial-temporal grounding.

\begin{table}[h]
\centering
\caption{%
  \textbf{Action conditioning ablation}: AdaLN vs.\ cross-attention across four environments
  spanning low and high action dimensionality.
  {\color{green!60!black}$\uparrow$} means clearly better than AdaLN.
}
\label{tab:abl_action_cond}
\resizebox{\linewidth}{!}{%
\begin{tabular}{l cccc cccc cccc cccc}
\toprule
&
\multicolumn{4}{c}{\textbf{Push Cube} ($d_a{=}2$)} &
\multicolumn{4}{c}{\textbf{Push Rope} ($d_a{=}2$)} &
\multicolumn{4}{c}{\textbf{Robot Arm} ($d_a{=}7$)} &
\multicolumn{4}{c}{\textbf{Cloth Move} ($d_a{=}8$)} \\
\cmidrule(lr){2-5}\cmidrule(lr){6-9}\cmidrule(lr){10-13}\cmidrule(lr){14-17}
\textbf{Method} &
  \multicolumn{2}{c}{InD} & \multicolumn{2}{c}{OoD} &
  \multicolumn{2}{c}{InD} & \multicolumn{2}{c}{OoD} &
  \multicolumn{2}{c}{InD} & \multicolumn{2}{c}{OoD} &
  \multicolumn{2}{c}{InD} & \multicolumn{2}{c}{OoD} \\
\cmidrule(lr){2-3}\cmidrule(lr){4-5}
\cmidrule(lr){6-7}\cmidrule(lr){8-9}
\cmidrule(lr){10-11}\cmidrule(lr){12-13}
\cmidrule(lr){14-15}\cmidrule(lr){16-17}
& MSE & PSNR & MSE & PSNR
  & MSE & PSNR & MSE & PSNR
  & MSE & PSNR & MSE & PSNR
  & MSE & PSNR & MSE & PSNR \\
\midrule
AdaLN (ours)
  & 2.919 & 25.35 & 2.950 & 25.30
  & 0.214 & 36.70 & 0.329 & 34.83
  & 1.434 & 28.43 & 6.559 & 21.83
  & \phantom{0}9.393 & 20.27 & \phantom{0}5.464 & 22.62 \\
Cross-Attn
  & 3.105 & 25.08 & 3.033 & 25.18
  & 0.216 & 36.65 & 0.334 & 34.77
  & 0.691{\color{green!60!black}$\uparrow$} & 31.61{\color{green!60!black}$\uparrow$} & 4.596{\color{green!60!black}$\uparrow$} & 23.38{\color{green!60!black}$\uparrow$}
  & 11.512 & 19.39 & \phantom{0}4.713{\color{green!60!black}$\uparrow$} & 23.27{\color{green!60!black}$\uparrow$} \\
\bottomrule
\end{tabular}}
\end{table}

%% ── 3. Latent-Space Formulation ───────────────────────────────────────────────
\finding{\textbf{Latent-Space Choice.} A temporally-aware causal VAE outperforms frame-independent VAE}
The Wan~2.1 causal VAE~\cite{wan21github} applies $4\times$ temporal compression, coupling consecutive frames
in the latent space.
We ablate this against a frame-independent image VAE (FLUX VAE~\cite{flux2024}),
which encodes each frame independently ($1\times$ temporal compression).
Table~\ref{tab:abl_vae} reports results on Pour Water and Robot Arm.
WanVAE outperforms FluxVAE on both InD and OoD scenarios, indicating that temporally-aware latent representations
are beneficial even for highly stochastic particle dynamics.

\begin{table}[t]
\centering
\caption{
\textbf{Latent-space formulation ablation.}
We compare a temporally-aware causal video VAE (Wan~2.1, $4\times$ temporal compression)
with a frame-independent image VAE (FLUX, $1\times$ temporal compression) on
\texttt{Pour Water} and \texttt{Robot Arm}.
}
\label{tab:abl_vae}
\small
\setlength{\tabcolsep}{4.5pt}
\begin{tabular}{l c cc cc cc cc}
\toprule
\multirow{2}{*}{\textbf{VAE}} &
\multirow{2}{*}{\textbf{Temp.}} &
\multicolumn{4}{c}{\textbf{Pour Water}} &
\multicolumn{4}{c}{\textbf{Robot Arm}} \\
\cmidrule(lr){3-6}\cmidrule(lr){7-10}
& &
\multicolumn{2}{c}{\textbf{InD}} &
\multicolumn{2}{c}{\textbf{OoD}} &
\multicolumn{2}{c}{\textbf{InD}} &
\multicolumn{2}{c}{\textbf{OoD}} \\
\cmidrule(lr){3-4}\cmidrule(lr){5-6}
\cmidrule(lr){7-8}\cmidrule(lr){9-10}
& &
MSE$\downarrow$ & PSNR$\uparrow$ &
MSE$\downarrow$ & PSNR$\uparrow$ &
MSE$\downarrow$ & PSNR$\uparrow$ &
MSE$\downarrow$ & PSNR$\uparrow$ \\
\midrule
FLUX VAE & $1\times$
& 3.241 & 24.89
& 3.895 & 24.09
& 2.314 & 26.50
& 10.143 & 20.07 \\
WanVAE (ours) & $4\times$
& \textbf{2.630} & \textbf{25.80}
& \textbf{3.488} & \textbf{24.57}
& \textbf{1.434} & \textbf{28.43}
& \textbf{6.559} & \textbf{21.83} \\
\bottomrule
\end{tabular}
\vspace{-0.1cm}
\end{table}

%% ── 4. Training Data Scaling ──────────────────────────────────────────────────
\finding{\textbf{Data Efficiency.} Data efficiency is governed by geometric variability: Pour Water degrades
gracefully with less data due to its single repeatable pouring motion, while
Push Cube degrades sharply as it requires diverse trajectory coverage over a wide workspace.}
We train DiT-S on Push Cube and Pour Water using 100\%, 50\%, and 25\% of the
available training trajectories (Table~\ref{tab:abl_data}).
Both environments show  degradation as data is reduced.
Push Cube degrades sharply,
indicating that diverse cube configurations and push directions require broad
trajectory coverage to generalize.
Pour Water is substantially more data-efficient: at 50\% data the InD drop is
only $0.26$\,dB, and even at 25\% it retains reasonable performance ($-1.69$\,dB),
likely because its dynamics are governed by a single repeatable pouring motion
with limited geometric variability.

\begin{table}[h]
\centering
\caption{%
  \textbf{Training data scaling ablation} on Push Cube and Pour Water.
  Models trained on 100\%, 50\%, and 25\% of available trajectories.
}
\label{tab:abl_data}
\small
\setlength{\tabcolsep}{5pt}
\begin{tabular}{lc cccc cccc}
\toprule
& &
\multicolumn{4}{c}{\textbf{Push Cube}} &
\multicolumn{4}{c}{\textbf{Pour Water}} \\
\cmidrule(lr){3-6}\cmidrule(lr){7-10}
\textbf{Data} & \textbf{Trajs} &
  \multicolumn{2}{c}{InD} & \multicolumn{2}{c}{OoD} &
  \multicolumn{2}{c}{InD} & \multicolumn{2}{c}{OoD} \\
\cmidrule(lr){3-4}\cmidrule(lr){5-6}\cmidrule(lr){7-8}\cmidrule(lr){9-10}
& & MSE$\downarrow$ & PSNR$\uparrow$ & MSE$\downarrow$ & PSNR$\uparrow$
  & MSE$\downarrow$ & PSNR$\uparrow$ & MSE$\downarrow$ & PSNR$\uparrow$ \\
\midrule
100\% (full) & 1987 / 1000 & 2.919 & 25.35 & 2.950 & 25.30 & 2.630 & 25.80 & 3.488 & 24.57 \\
50\%         &  994 / 500  & 3.724 & 24.29 & 3.981 & 24.00 & 2.793 & 25.54 & 3.623 & 24.41 \\
25\%         &  497 / 250  & 5.318 & 22.74 & 5.891 & 22.30 & 3.882 & 24.11 & 4.689 & 23.29 \\
\bottomrule
\end{tabular}
\end{table}

%% ── 5. Action Dimensionality ──────────────────────────────────────────────────
\finding{\textbf{Action Dimensionality.} Higher-dimensional actions make prediction harder, but can improve OoD generalization when they provide more informative control signals and richer observation.}
Table~\ref{tab:abl_actdim} compares action-space variants for Cloth Move and Push Cube. For Cloth Move, we use the full action space by allowing the two grippers to move independently, providing more detailed control over the deformable object. This substantially improves OoD MSE, suggesting that richer action signals help the model infer two-arm cloth dynamics. In contrast, for Push Cube, adding a second pusher increases interaction complexity without providing the same informative benefit, resulting in higher MSE in both InD and OoD settings.

\section{Conclusion and Limitation}

We introduced ACWM-Phys, a benchmark spanning four physically diverse interaction regimes, and ACWM-DiT, a latent diffusion transformer baseline trained with flow matching. Our experiments show that current ACWMs achieve strong in-distribution fidelity but suffer substantial OoD degradation that correlates with physical complexity: rigid-body and kinematic tasks generalize relatively well, while deformable and particle-dynamics tasks expose larger gaps, suggesting that models still rely heavily on appearance statistics rather than internalizing physics. Ablations further show that larger models, temporally-aware VAEs, and richer action specifications improve OoD robustness, especially on harder high-dimensional tasks. We hope ACWM-Phys serves as a diagnostic tool for the community and encourages future work on architectures that explicitly represent physical structure.

\paragraph{Limitations.}
First, ACWM-DiT is not designed for real-time rendering. As a bidirectional diffusion model, it provides strong visual prediction quality but remains slow at inference. Future work may explore autoregressive diffusion models e.g. with diffusion forcing or self-forcing to support real-time world-model rollout.
Second, ACWM-Phys is built in simulation, which enables controlled OoD evaluation but does not fully capture the complexity of real-world physics, sensing, and robot interaction. Bridging this gap may require sim-to-real transfer, more realistic simulators, or human demonstration data collected in real environments.
\bibliography{main}

\begin{thebibliography}{10}

\bibitem{bagchi2026walk}
A.~Bagchi, Z.~Bao, H.~Bharadhwaj, Y.-X. Wang, P.~Tokmakov, and M.~Hebert.
\newblock Walk through paintings: Egocentric world models from internet priors.
\newblock {\em arXiv preprint arXiv:2601.15284}, 2026.

\bibitem{chen2026bridgev2w}
Y.~Chen, P.~Li, J.~Yang, K.~He, X.~Wu, Y.~Xu, K.~Wang, J.~Liu, N.~Liu, Y.~Huang, et~al.
\newblock Bridgev2w: Bridging video generation models to embodied world models via embodiment masks.
\newblock {\em arXiv preprint arXiv:2602.03793}, 2026.

\bibitem{guo2025ctrl}
Y.~Guo, L.~X. Shi, J.~Chen, and C.~Finn.
\newblock Ctrl-world: A controllable generative world model for robot manipulation.
\newblock {\em arXiv preprint arXiv:2510.10125}, 2025.

\bibitem{2018worldmodel}
D.~Ha and J.~Schmidhuber.
\newblock World models.
\newblock {\em arXiv preprint arXiv:1803.10122}, 2(3):440, 2018.

\bibitem{dreamer4}
D.~Hafner, W.~Yan, and T.~Lillicrap.
\newblock Training agents inside of scalable world models.
\newblock {\em arXiv preprint arXiv:2509.24527}, 2025.

\bibitem{ho2020ddpm}
J.~Ho, A.~Jain, and P.~Abbeel.
\newblock Denoising diffusion probabilistic models.
\newblock In {\em Advances in Neural Information Processing Systems}, volume~33, 2020.

\bibitem{ho2022video}
J.~Ho, T.~Salimans, A.~Gritsenko, W.~Chan, M.~Norouzi, and D.~J. Fleet.
\newblock Video diffusion models.
\newblock {\em Advances in neural information processing systems}, 35:8633--8646, 2022.

\bibitem{RelicWorldModel2025}
Y.~Hong, Y.~Mei, C.~Ge, Y.~Xu, Y.~Zhou, S.~Bi, Y.~Hold-Geoffroy, M.~Roberts, M.~Fisher, E.~Shechtman, K.~Sunkavalli, F.~Liu, Z.~Li, and H.~Tan.
\newblock Relic: Interactive video world models with long-horizon memory, 2025.

\bibitem{psnr}
A.~Hore and D.~Ziou.
\newblock Image quality metrics: Psnr vs. ssim.
\newblock In {\em 2010 20th international conference on pattern recognition}, pages 2366--2369. IEEE, 2010.

\bibitem{huang2025vid2world}
S.~Huang, J.~Wu, Q.~Zhou, S.~Miao, and M.~Long.
\newblock Vid2world: Crafting video diffusion models to interactive world models.
\newblock {\em arXiv preprint arXiv:2505.14357}, 2025.

\bibitem{huang2025selfforcing}
X.~Huang, Z.~Li, G.~He, M.~Zhou, and E.~Shechtman.
\newblock Self forcing: Bridging the train-test gap in autoregressive video diffusion.
\newblock {\em arXiv preprint arXiv:2506.08009}, 2025.

\bibitem{jiang2026wovr}
Z.~Jiang, S.~Zhou, Y.~Jiang, Z.~Huang, M.~Wei, Y.~Chen, T.~Zhou, Z.~Guo, H.~Lin, Q.~Zhang, et~al.
\newblock Wovr: World models as reliable simulators for post-training vla policies with rl.
\newblock {\em arXiv preprint arXiv:2602.13977}, 2026.

\bibitem{kang2024far}
B.~Kang, Y.~Yue, R.~Lu, Z.~Lin, Y.~Zhao, K.~Wang, G.~Huang, and J.~Feng.
\newblock How far is video generation from world model: A physical law perspective.
\newblock {\em arXiv preprint arXiv:2411.02385}, 2024.

\bibitem{karras2022edm}
T.~Karras, M.~Aittala, T.~Aila, and S.~Laine.
\newblock Elucidating the design space of diffusion-based generative models.
\newblock In {\em Advances in Neural Information Processing Systems}, 2022.

\bibitem{khazatsky2024droid}
A.~Khazatsky, K.~Pertsch, S.~Nair, A.~Balakrishna, S.~Dasari, S.~Karamcheti, S.~Nasiriany, M.~K. Srirama, L.~Y. Chen, K.~Ellis, et~al.
\newblock Droid: A large-scale in-the-wild robot manipulation dataset.
\newblock {\em arXiv preprint arXiv:2403.12945}, 2024.

\bibitem{flux2024}
B.~F. Labs.
\newblock Flux.
\newblock \url{https://github.com/black-forest-labs/flux}, 2024.

\bibitem{le2025gravity}
M.-Q. Le, Y.~Zhu, V.~Kalogeiton, and D.~Samaras.
\newblock What about gravity in video generation? post-training newton's laws with verifiable rewards.
\newblock {\em arXiv preprint arXiv:2512.00425}, 2025.

\bibitem{li2025pisa}
C.~Li, O.~Michel, X.~Pan, S.~Liu, M.~Roberts, and S.~Xie.
\newblock Pisa experiments: Exploring physics post-training for video diffusion models by watching stuff drop.
\newblock {\em arXiv preprint arXiv:2503.09595}, 2025.

\bibitem{li2018learning}
Y.~Li, J.~Wu, R.~Tedrake, J.~B. Tenenbaum, and A.~Torralba.
\newblock Learning particle dynamics for manipulating rigid bodies, deformable objects, and fluids.
\newblock {\em arXiv preprint arXiv:1810.01566}, 2018.

\bibitem{lipman2022flow}
Y.~Lipman, R.~T. Chen, H.~Ben-Hamu, M.~Nickel, and M.~Le.
\newblock Flow matching for generative modeling.
\newblock {\em arXiv preprint arXiv:2210.02747}, 2022.

\bibitem{liu2022flow}
X.~Liu, C.~Gong, and Q.~Liu.
\newblock Flow straight and fast: Learning to generate and transfer data with rectified flow.
\newblock {\em arXiv preprint arXiv:2209.03003}, 2022.

\bibitem{motamed2026generative}
S.~Motamed, L.~Culp, K.~Swersky, P.~Jaini, and R.~Geirhos.
\newblock Do generative video models understand physical principles?
\newblock In {\em Proceedings of the IEEE/CVF Winter Conference on Applications of Computer Vision}, pages 948--958, 2026.

\bibitem{parker2025genie}
J.~Parker-Holder and S.~Fruchter.
\newblock Genie 3: A new frontier for world models.
\newblock {\em URL https://deepmind. google/discover/blog/genie-3-a-new-frontier-for-world-models/. Blog post}, 2025.

\bibitem{peebles2023scalable}
W.~Peebles and S.~Xie.
\newblock Scalable diffusion models with transformers.
\newblock In {\em Proceedings of the IEEE/CVF international conference on computer vision}, pages 4195--4205, 2023.

\bibitem{savva2026solaris}
G.~Savva, O.~Michel, D.~Lu, S.~Waiwitlikhit, T.~Meehan, D.~Mishra, S.~Poddar, J.~Lu, and S.~Xie.
\newblock Solaris: Building a multiplayer video world model in minecraft.
\newblock {\em arXiv preprint arXiv:2602.22208}, 2026.

\bibitem{recondataset}
D.~Shah, B.~Eysenbach, N.~Rhinehart, and S.~Levine.
\newblock Rapid exploration for open-world navigation with latent goal models.
\newblock In {\em 5th Annual Conference on Robot Learning}, 2021.

\bibitem{sun2025worldplay}
W.~Sun, H.~Zhang, H.~Wang, J.~Wu, Z.~Wang, Z.~Wang, Y.~Wang, J.~Zhang, T.~Wang, and C.~Guo.
\newblock Worldplay: Towards long-term geometric consistency for real-time interactive world modeling.
\newblock {\em arXiv preprint arXiv:2512.14614}, 2025.

\bibitem{todorov2012mujoco}
E.~Todorov, T.~Erez, and Y.~Tassa.
\newblock Mujoco: A physics engine for model-based control.
\newblock In {\em 2012 IEEE/RSJ International Conference on Intelligent Robots and Systems}, pages 5026--5033. IEEE, 2012.

\bibitem{wan2025wan}
T.~Wan et~al.
\newblock Open and advanced large-scale video generative models.
\newblock {\em arXiv preprint arXiv:2503.20314}, 2025.

\bibitem{wan21github}
{Wan-Video Team}.
\newblock {Wan2.1}: Open video foundation models.
\newblock {GitHub} repository, 2025.
\newblock Technical report and weights; project page details evolving.

\bibitem{wang2025wisa}
J.~Wang, A.~Ma, K.~Cao, J.~Zheng, J.~Feng, Z.~Zhang, W.~Pang, and X.~Liang.
\newblock Wisa: World simulator assistant for physics-aware text-to-video generation.
\newblock In {\em Advances in Neural Information Processing Systems}, 2025.

\bibitem{wang2025prophy}
Z.~Wang, P.~Hu, J.~Wang, T.~J. Zhang, Y.~Cheng, L.~Chen, Y.~Yan, Z.~Jiang, H.~Li, and X.~Liang.
\newblock Prophy: Progressive physical alignment for dynamic world simulation.
\newblock {\em arXiv preprint arXiv:2512.05564}, 2025.

\bibitem{wang2025videoverse}
Z.~Wang, X.~Wei, B.~Li, Z.~Guo, J.~Zhang, H.~Wei, K.~Wang, and L.~Zhang.
\newblock Videoverse: How far is your t2v generator from a world model?
\newblock {\em arXiv preprint arXiv:2510.08398}, 2025.

\bibitem{yang2024cogvideox}
Z.~Yang et~al.
\newblock {CogVideoX}: Text-to-video diffusion models with an expert transformer.
\newblock {\em arXiv preprint arXiv:2408.06072}, 2024.

\bibitem{ye2026world}
S.~Ye, Y.~Ge, K.~Zheng, S.~Gao, S.~Yu, G.~Kurian, S.~Indupuru, Y.~L. Tan, C.~Zhu, J.~Xiang, et~al.
\newblock World action models are zero-shot policies.
\newblock {\em arXiv preprint arXiv:2602.15922}, 2026.

\bibitem{yuan2026newtongen}
Y.~Yuan, X.~Wang, T.~Wickremasinghe, Z.~Nadir, B.~Ma, and S.~H. Chan.
\newblock Newtongen: Physics-consistent and controllable text-to-video generation via neural newtonian dynamics.
\newblock In {\em International Conference on Learning Representations}, 2026.

\bibitem{zhang2025morpheus}
C.~Zhang, D.~Cherniavskii, A.~Tragoudaras, A.~Vozikis, T.~Nijdam, D.~W. Prinzhorn, M.~Bodracska, N.~Sebe, A.~Zadaianchuk, and E.~Gavves.
\newblock Morpheus: Benchmarking physical reasoning of video generative models with real physical experiments.
\newblock {\em arXiv preprint arXiv:2504.02918}, 2025.

\bibitem{zhang2025thinkdiffusellmsguidedphysicsaware}
K.~Zhang, C.~Xiao, Y.~Mei, J.~Xu, and V.~M. Patel.
\newblock Think before you diffuse: Llms-guided physics-aware video generation, 2025.

\bibitem{physinone}
S.~Zhou, H.~Wang, H.~Cheng, J.~Li, D.~Wang, J.~Jiang, Y.~Jin, J.~Huang, S.~Mao, S.~Liu, Y.~Yang, H.~Song, S.~Wei, Z.~Zhang, P.~Huang, S.~Liu, Z.~Hao, H.~Li, Y.~Li, W.~Zhou, Z.~Zhao, Z.~He, H.~Wen, S.~Huang, P.~Yun, B.~Cheng, P.~K. Fu, W.~K. Lai, J.~Chen, K.~Wang, Z.~Sun, Z.~Li, H.~Hu, D.~Zhang, C.~H. Yuen, B.~Wang, Z.~Wang, C.~Zou, and B.~Yang.
\newblock Physinone: Visual physics learning and reasoning in one suite, 2026.

\end{thebibliography}
\bibliographystyle{abbrv}

\newpage
\appendix
\section{Appendix}

\subsection{Details of ACWM-DiT}\label{sec:app-model-structure}
\paragraph{Overall Architecture}

ACWM-DiT (Figure~\ref{fig:model_arch}) is a latent video diffusion transformer trained
with flow matching to predict future observation trajectories conditioned on past
observations and a sequence of actions.

\textbf{Video VAE.}
Following Wan~2.1~\cite{wan2025wan}, we use a pretrained causal VAE that compresses a
video $\mathbf{o}_{1:T}\in\mathbb{R}^{T\times 3\times H\times W}$ into latent tokens
$\mathbf{z}\in\mathbb{R}^{T_l\times\frac{H}{8}\times\frac{W}{8}\times 16}$, applying
$8\times$ spatial compression, $4\times$ temporal compression, and a 16-channel latent
space.  All VAE weights are frozen throughout training.

\textbf{Video DiT Backbone.}
The denoiser is a bidirectional (non-causal) transformer that operates on
spatiotemporally patchified latents (spatial patch size~2).  Each of the $N$ transformer
blocks consists of a spatial self-attention layer (attending over the
$\frac{H}{16}\times\frac{W}{16}$ spatial patches within each frame) followed by a
temporal self-attention layer attending across $T_l$ latent timesteps at each spatial
location.  Rotary Position Embeddings (RoPE) are used in both layers.
Observation-conditioning frames are passed at noise level $\sigma{=}0$ while predicted
frames are noised at the current diffusion step $\sigma$; this allows the model to
attend bidirectionally over both clean context and noisy future tokens.
Conditioning on the diffusion timestep and actions is applied through Adaptive Layer
Normalization (AdaLN), which uses the joint signal $\mathbf{c}$ to produce per-step
scale and shift parameters $(\gamma,\beta,\alpha)$ at every block.
We study two model scales: \textbf{DiT-S} (hidden dim~768, 10~layers, 12~heads,
${\approx}200$M parameters) and \textbf{DiT-M} (hidden dim~1024, 16~layers, 16~heads,
${\approx}600$M parameters).

\textbf{Flow Matching.}
Training follows the latent-space flow matching objective defined in
Section~\ref{sec:background} with a linear interpolation path and shift
parameter~$s{=}5.0$.  Training losses are weighted by a Gaussian envelope centered at
diffusion step~500 to focus supervision on intermediate noise levels.  We use 1000
training steps and 50 inference denoising steps.

\paragraph{Action Conditioning Module}

Actions are integrated via a dedicated \emph{ActionEmbedder} module that maps the
pixel-rate action sequence to the latent temporal resolution in two stages.
First, an MLP (Linear$\to$SiLU$\to$Linear) projects each action vector from its
environment-specific dimension $d_a$ to the DiT hidden dimension~$d$.
Second, a 1D strided convolution (kernel size~3, stride~$r{=}4$, matching the
VAE's $4\times$ temporal compression factor) downsamples the sequence from
$1 + r(T_l{-}1)$ pixel-rate steps to $T_l$ latent-rate tokens, producing
$\hat{\mathbf{a}}\in\mathbb{R}^{T_l\times d}$.

The action embedding is summed with the per-step timestep embedding to form the
joint conditioning signal
$\mathbf{c} = \mathbf{c}_t + \hat{\mathbf{a}} \;\in\; \mathbb{R}^{T_l\times d},$
which is then broadcast into every DiT block via AdaLN, modulating the scale and
shift applied after each layer normalization.
This design couples action information with the diffusion timestep and propagates it
uniformly across all spatial and temporal attention operations without requiring
additional cross-attention layers or token-sequence modifications.
For the standard training setting we use no action dropout ($p_{\text{drop}}{=}0$);
ablations on classifier-free guidance are left to future work.

\subsection{Out-of-Distribution Split Design}\label{app:ood-design}

Each ACWM-Phys environment defines a physically motivated OoD split that targets a specific axis of generalization rather than using random perturbations. Table~\ref{tab:ood_design} summarizes the shift type and parameter range for each environment. All OoD trajectories are generated by the same simulator as the training data, but with configurations explicitly excluded from the training distribution.

\paragraph{Rigid-Body.}
For \texttt{Push Cube}, training covers a central workspace region, while OoD episodes place the cube initial positions and push targets near table corners or edges, requiring spatial extrapolation of rigid-body dynamics. We also include a harder variant, \texttt{push\_cube\_4cube}, with four or more cubes to test generalization to unseen object counts.
For \texttt{Stack Cube}, the target placement direction defines the shift: training uses a subset of directions, while OoD episodes require stacking toward held-out cardinal directions, recorded by the \texttt{ood\_label} field.

\paragraph{Deformable.}
For \texttt{Push Rope}, the OoD shift is rope length, which affects stiffness and deformability. Training uses rope lengths in $[2.0, 2.8]$\,m, while OoD episodes fix the length to $3.1$\,m, producing qualitatively different deformation dynamics.
For \texttt{Cloth Move}, OoD episodes change the cloth size and initial configuration, placing the cloth in spatial arrangements not observed during training.

\paragraph{Particle.}
For \texttt{Push Sand}, OoD episodes use different random seeds from training/InD testing, resulting in unseen initial particle layouts and density configurations.
For \texttt{Pour Water}, the shift is the water quantity: InD episodes cover a nominal water-level range, while OoD episodes use substantially lower or higher volumes, testing extrapolation of fluid dynamics.

\paragraph{Kinematics.}
For \texttt{Robot Arm}, OoD episodes require reaching targets in an expanded workspace. The joint-angle action range expands from $[-0.95, 0.58]$ in InD to $[-1.15, 0.88]$ in OoD, inducing more extreme arm configurations than those seen during training.
For \texttt{Reacher}, OoD goals are sampled from corner sectors of the reachable space excluded during training, with the action range expanding from $[-3.3, 3.5]$ to $[-3.7, 4.2]$\,rad.

\subsection{Masked-MSE (M-MSE)}\label{app:mmse}

Standard MSE treats all pixels equally and can therefore be dominated by static background regions, especially in environments where only a small portion of the scene is physically dynamic. To better focus evaluation on moving regions, we introduce \textbf{Masked-MSE (M-MSE)}, a motion-aware weighted mean squared error.

Given a ground-truth video $\mathbf{o}_{1:T}$ and a predicted video $\hat{\mathbf{o}}_{1:T}$, we first compute a per-pixel motion map using only the ground-truth video:
\[
m_{h,w} = \max_{t \in [1,T],\, c} \left|\mathbf{o}_{t,c,h,w} - \mathbf{o}_{1,c,h,w}\right|,
\]
which measures the maximum absolute deviation of each pixel from the first frame across all timesteps and channels. We then define a soft per-pixel weight:
\[
w_{h,w} = 0.01 + m_{h,w},
\]
where the small floor value prevents zero weights for perfectly static pixels while remaining negligible compared to typical motion magnitudes. In practice, the weight is therefore dominated by $m_{h,w}$ and assigns very small weights to static background pixels.

M-MSE is defined as the resulting weighted mean squared error:
\[
\text{M-MSE} =
\frac{
\sum_{t,c,h,w} w_{h,w}
\left(\hat{\mathbf{o}}_{t,c,h,w} - \mathbf{o}_{t,c,h,w}\right)^2
}{
\sum_{t,c,h,w} w_{h,w}
}.
\]
Because $w_{h,w}$ is proportional to the amount of motion at each pixel, errors in dynamic foreground regions are strongly up-weighted, while static background pixels contribute only minimally. This makes M-MSE more sensitive to physically meaningful prediction failures that may be under-emphasized by standard MSE.

% in preamble
% \usepackage{booktabs}
% \usepackage{threeparttable}
% \usepackage{pifont}
% \newcommand{\cmark}{\ding{51}}
% \newcommand{\xmark}{\ding{55}}

\begin{table*}[t]
\centering
\small
\setlength{\tabcolsep}{5pt}
\renewcommand{\arraystretch}{1.3}
\begin{threeparttable}
\resizebox{\textwidth}{!}{
\begin{tabular}{l l c c c c}
\toprule
\textbf{Benchmark} & \textbf{Type} & \textbf{Action-conditioned} & \textbf{Rich Physics} & \textbf{Data Scale} & \textbf{ID+OOD} \\
\midrule
Solaris~\cite{savva2026solaris}
& Multi-Player Minecraft
& \cmark
& \xmark
& 9k videos
& \xmark \\

Dreamer4~\cite{dreamer4}
& Minecraft
& \cmark
& \xmark
& 2.5K hour gameplay
& \xmark \\

DROID~\cite{khazatsky2024droid}
& Expert manipulation
& \cmark
& \xmark
& 76K trajectories
& \xmark \\

RECON~\cite{recondataset}
& Egocentric navigation
& \cmark
& \xmark
& 5K trajectories
& \xmark \\

Physics-IQ~\cite{motamed2026generative}
& Passive physics evaluation
& \xmark
& \cmark
& 396 videos
& \xmark \\

WISA~\cite{wang2025wisa}
& Passive physics evaluation
& \xmark
& \cmark
& 32K videos
& \xmark \\

PhysInOne~\cite{physinone}
& Synthetic physics interaction
& \xmark
& \cmark
& 2M videos
& \xmark \\

\midrule
\textbf{ACWM-Phys (ours)}
& \textbf{Synthetic physics interaction}
& \textbf{\cmark}
& \textbf{\cmark}
& \textbf{15k videos}
& \textbf{\cmark} \\
\bottomrule
\end{tabular}
}
\end{threeparttable}
\caption{\textbf{Comparison of ACWM-Phys with existing action-conditioned and physics-related benchmarks.} Benchmarks used for training action-conditioned world models generally lack rich and diverse physical interactions, whereas existing physics-focused benchmarks are typically not action-conditioned. Also, most prior benchmarks do not include both in-distribution and out-of-distribution evaluations}
\label{tab:benchmark_comparison}
\end{table*}

\begin{table}[h]
\centering
\caption{%
  \textbf{Action dimensionality ablation.}
  Cloth Move compares the reduced 3-DoF shared action against the full 8-DoF
  per-arm action space.
  Push Cube compares a single pusher ($d_a{=}2$) against two independent
  pushers ($d_a{=}4$).
}
\label{tab:abl_actdim}
\small
\setlength{\tabcolsep}{5pt}
\begin{tabular}{llc cccc}
\toprule
\textbf{Environment} & \textbf{Action Space} & $d_a$ &
  \multicolumn{2}{c}{\textbf{InD}} & \multicolumn{2}{c}{\textbf{OoD}} \\
\cmidrule(lr){4-5}\cmidrule(lr){6-7}
& & & MSE$\downarrow$ & PSNR$\uparrow$ & MSE$\downarrow$ & PSNR$\uparrow$ \\
\midrule
\multirow{2}{*}{Cloth Move}
  & Reduced (shared $\Delta xyz$) & 3 & 10.667 & 19.72 & 23.820 & 16.23 \\
  & Full (per-arm $\Delta$pose + grasp) & 8 & 12.861 & 18.91 & \phantom{0}6.922 & 21.60 \\
\midrule
\multirow{2}{*}{Push Cube}
  & Single pusher & 2 & 2.919 & 25.35 & 2.950 & 25.30 \\
  & Two pushers   & 4 & 6.034 & 22.19 & 9.997 & 20.00 \\
\bottomrule
\end{tabular}
\end{table}

\begin{figure}[h]
  \centering
  \definecolor{arblue}{RGB}{37,99,235}
\definecolor{arred}{RGB}{220,38,38}
{\setlength{\tabcolsep}{2pt}%
\renewcommand{\arraystretch}{0.9}%
\scriptsize%
\begin{tabular}{@{}r@{\hspace{4pt}}
    c@{\hspace{1pt}}c
    @{\hspace{10pt}}
    c@{\hspace{1pt}}c
    @{}}
  %% env headers
  & \multicolumn{2}{c}{\textbf{Pour Water}}
  & \multicolumn{2}{c}{\textbf{Push Cube}} \\[1pt]
  %% GT row
  \rotatebox{90}{\textbf{GT}}
    & \includegraphics[width=0.225\linewidth]{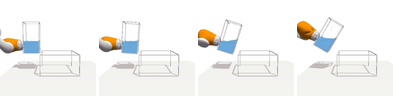}
    & \includegraphics[width=0.225\linewidth]{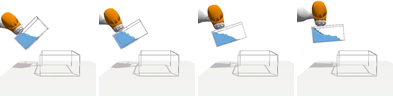}
    & \includegraphics[width=0.225\linewidth]{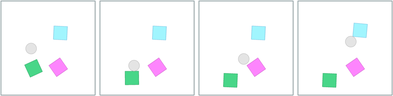}
    & \includegraphics[width=0.225\linewidth]{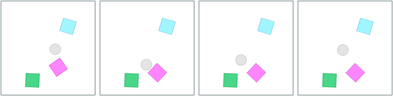} \\[0.5pt]
  %% Pred row
  \rotatebox{90}{\textbf{Pred}}
    & \includegraphics[width=0.225\linewidth]{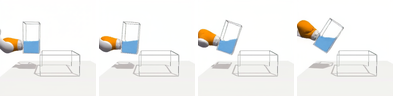}
    & \includegraphics[width=0.225\linewidth]{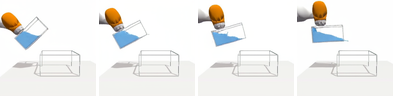}
    & \includegraphics[width=0.225\linewidth]{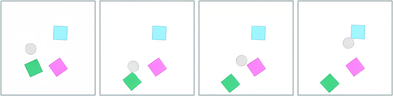}
    & \includegraphics[width=0.225\linewidth]{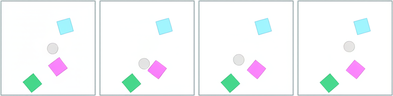} \\[1pt]
  %% time-range underbrace at the bottom
  & $\underbrace{\hspace{0.225\linewidth}}_{\textcolor{arblue}{\boldsymbol{1{\to}37}}}$
  & $\underbrace{\hspace{0.225\linewidth}}_{\textcolor{arred}{\boldsymbol{37{\to}74}}}$
  & $\underbrace{\hspace{0.225\linewidth}}_{\textcolor{arblue}{\boldsymbol{1{\to}37}}}$
  & $\underbrace{\hspace{0.225\linewidth}}_{\textcolor{arred}{\boldsymbol{37{\to}66}}}$ \\
\end{tabular}}

  \caption{%
    \textbf{Auto-regressive Generation.}
    The model generates frames $1{\to}37$ (blue) conditioned on the first frame,
    then generates frames $37{\to}T$ (red) conditioned on the last predicted frame
    of the first window.
    GT (top) and predicted (bottom) frames at four evenly-spaced timesteps per window.
  }
  \label{fig:ar_generation}
\end{figure}
%%%%%%%%%%%%%%%%%%%%%%%%%%%%%%%%%%%%%%%%%%%%%%%%%%%%%%%%%%%%

\subsection{Dataset Statistics and Action Space Definitions}
\label{app:dataset_stats}

Tables~\ref{tab:dataset} and~\ref{tab:action_space} summarize the dataset sizes,
action dimensionalities, and trajectory horizons across all eight ACWM-Phys environments,
together with the detailed semantics of each action space.

% Dataset statistics table for ACWM-Phys
\begin{table}[t]
\centering
\caption{%
  \textbf{ACWM-Phys dataset statistics.}
  Each environment provides in-distribution (InD) and out-of-distribution (OoD) test splits
  with controlled distribution shifts. Action dim refers to the dimensionality of the
  action vector fed to ACWM-DiT; full action definitions are listed in Table~\ref{tab:action_space}.
}
\label{tab:dataset}
\small
\setlength{\tabcolsep}{5pt}
\begin{threeparttable}
\begin{tabular}{llccccc}
\toprule
\textbf{Category} & \textbf{Environment} & \textbf{Train} & \textbf{InD Test} & \textbf{OoD Test} & \textbf{Act.\ Dim} & \textbf{Horizon} \\
\midrule
\multirow{2}{*}{Rigid Dynamics}
  & Push Cube   & 1,987 & 100 & 100 & 2 & $\sim$66 \\
  & Stack Cube  & 1,987 & 100 & 100 & 7 & $\sim$57 \\
\midrule
\multirow{2}{*}{Deformable}
  & Push Rope       & 1,987 & 100 & 100 & 2 & 64 \\
  & Cloth Move$^*$  & 1,987 & 100 & 100 & 3 & 64 \\
\midrule
\multirow{2}{*}{Particle}
  & Push Sand   & 1,784 & 100 & 100 & 7 & 64 \\
  & Pour Water  & 1,000 &  50 &  50 & 4 & 300 \\
\midrule
\multirow{2}{*}{Kinematics}
  & Robot Arm   & 1,987 & 100 & 100 & 7 & 64 \\
  & Reacher     & 1,987 & 100 & 100 & 2 & 50  \\
\midrule
\textbf{Total}    & 8 environments       & \textbf{14,706} & \textbf{750} & \textbf{750} & 2--7 & 50--300 \\
\bottomrule
\multicolumn{7}{l}{\footnotesize $^*$Cloth Move uses reduced 3-dim actions; 8-dim variant used in action-dimension ablation.}
\end{tabular}
\end{threeparttable}
\end{table}

% Action space summary table
\begin{table}[t]
\centering
\caption{%
  \textbf{Action space definitions across ACWM-Phys environments.}
  Actions are either \emph{absolute} target states or \emph{delta} (incremental) commands.
}
\label{tab:action_space}
\small
\setlength{\tabcolsep}{5pt}
\begin{tabular}{llcl}
\toprule
\textbf{Environment} & \textbf{Type} & \textbf{Dim} & \textbf{Semantics} \\
\midrule
Push Cube  & Absolute position  & 2 & Target $(x,y)$ pusher position on workspace \\
Stack Cube & Delta EE pose      & 7 & $(\Delta x,\Delta y,\Delta z,\Delta r_x,\Delta r_y,\Delta r_z)$ + gripper open/close \\
Push Rope  & Delta position     & 2 & 2D horizontal pole displacement $(\Delta x, \Delta z)$ \\
Cloth Move & Delta position      & 3 & $(\Delta x,\Delta y,\Delta z)$ dual-arm displacement$^*$ \\
Push Sand  & Delta board pose   & 7 & Board 3D position + 4D orientation delta \\
Pour Water & Delta cup pose     & 4 & $(\Delta x,\Delta y,\Delta z,\Delta\theta)$ cup tilt delta \\
Robot Arm  & Delta joint angles & 7 & Per-joint angle delta $\Delta\mathbf{q}\in\mathbb{R}^7$ \\
Reacher    & Joint torques      & 2 & Torques $(\tau_1,\tau_2)$ applied to two links \\
\bottomrule
\end{tabular}
\end{table}

%% ─────────────────────────────────────────────────────────────────────────────
\subsection{Metrics vs.\ Diffusion Steps}
\label{app:steps_curves}

\begin{figure*}[!h]
  \centering
  \includegraphics[width=\linewidth]{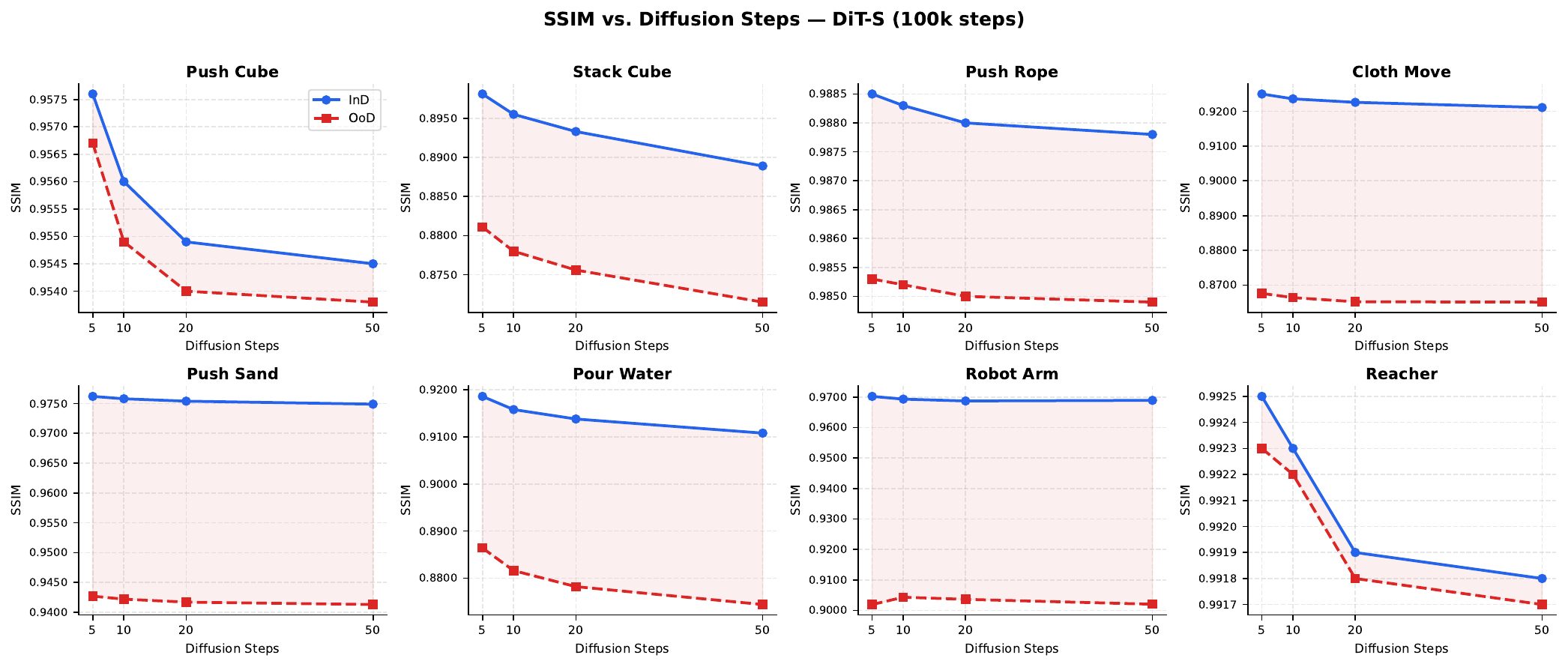}
  \caption{%
    \textbf{SSIM vs.\ diffusion steps} for ACWM-DiT-S (100k training steps).
    Blue circles: InD test; red squares: OoD test.
    Higher SSIM is better ($\uparrow$).
  }
  \label{fig:ssim_vs_steps}
\end{figure*}

\begin{figure*}[!h]
  \centering
  \includegraphics[width=\linewidth]{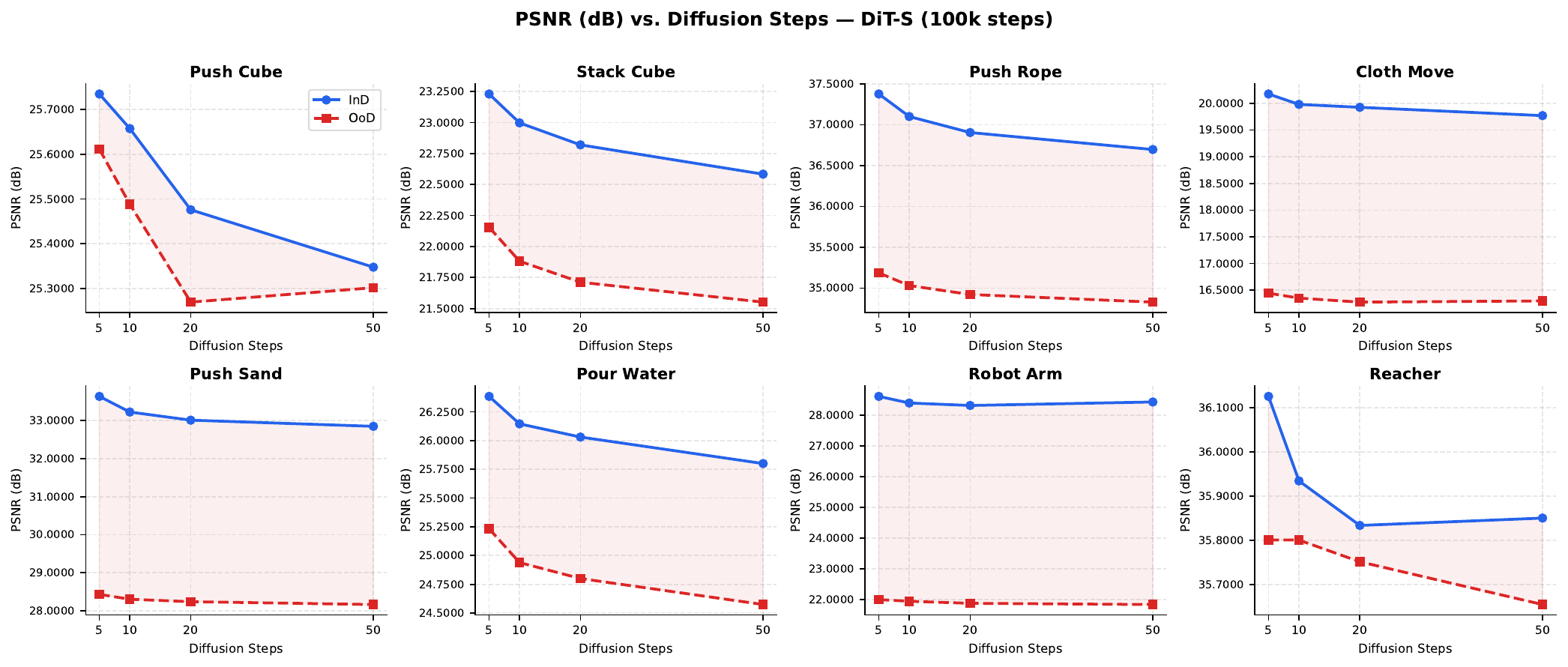}
  \caption{%
    \textbf{PSNR vs.\ diffusion steps} for ACWM-DiT-S (100k training steps).
    Blue circles: InD test; red squares: OoD test.
    Higher PSNR is better ($\uparrow$).
  }
  \label{fig:psnr_vs_steps}
\end{figure*}

Figures~\ref{fig:ssim_vs_steps} and~\ref{fig:psnr_vs_steps} show SSIM and PSNR
as a function of the number of inference diffusion steps across all eight ACWM-Phys
environments.  In general, performance saturates quickly: gains from 5 to 50 steps
are marginal for most tasks, suggesting that 5--10 steps suffice at inference time.

%% ─────────────────────────────────────────────────────────────────────────────
\subsection{Dataset Visualizations}
\label{app:dataset_viz}

\begin{figure}[!h]
  \centering
  % Two-column layout at 0.455\linewidth each (~0.8 total) — fits on one page.
% Left: Rigid-body + Deformable  |  Right: Particle + Kinematics
{\setlength{\tabcolsep}{1pt}%
\renewcommand{\arraystretch}{0.80}%
\scriptsize%
\begin{center}
\begin{minipage}[t]{0.455\linewidth}
  \begin{tabular}{@{}r@{\hspace{2pt}}>{\centering\arraybackslash}p{\dimexpr\linewidth-14pt\relax}@{}}
    %% Rigid-body ─────────────────────────────────────────────────────────
    \multicolumn{2}{@{}l@{}}{\textbf{\textsc{Push Cube}}} \\[0.5pt]
    \rotatebox{90}{\textbf{InD}} & \includegraphics[width=\linewidth]{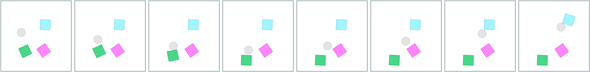} \\[0.3pt]
    \rotatebox{90}{\textbf{OoD}} & \includegraphics[width=\linewidth]{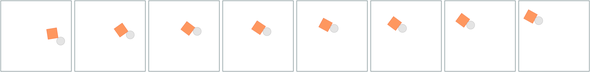} \\[3pt]
    \multicolumn{2}{@{}l@{}}{\textbf{\textsc{Stack Cube}}} \\[0.5pt]
    \rotatebox{90}{\textbf{InD}} & \includegraphics[width=\linewidth]{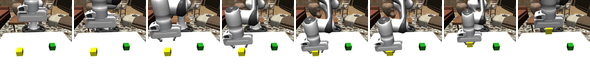} \\[0.3pt]
    \rotatebox{90}{\textbf{OoD}} & \includegraphics[width=\linewidth]{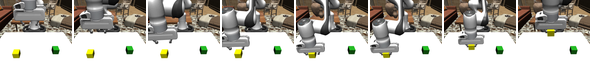} \\[3pt]
    %% Deformable ──────────────────────────────────────────────────────────
    \multicolumn{2}{@{}l@{}}{\textbf{\textsc{Push Rope}}} \\[0.5pt]
    \rotatebox{90}{\textbf{InD}} & \includegraphics[width=\linewidth]{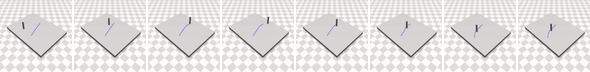} \\[0.3pt]
    \rotatebox{90}{\textbf{OoD}} & \includegraphics[width=\linewidth]{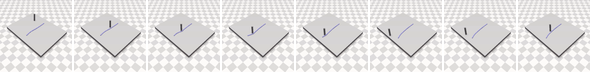} \\[3pt]
    \multicolumn{2}{@{}l@{}}{\textbf{\textsc{Cloth Move}}} \\[0.5pt]
    \rotatebox{90}{\textbf{InD}} & \includegraphics[width=\linewidth]{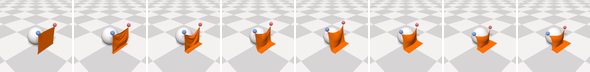} \\[0.3pt]
    \rotatebox{90}{\textbf{OoD}} & \includegraphics[width=\linewidth]{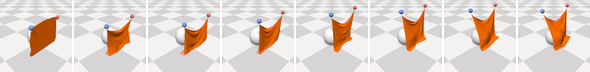} \\
  \end{tabular}
\end{minipage}%
\hspace{0.04\linewidth}%
\begin{minipage}[t]{0.455\linewidth}
  \begin{tabular}{@{}r@{\hspace{2pt}}>{\centering\arraybackslash}p{\dimexpr\linewidth-14pt\relax}@{}}
    %% Particle ────────────────────────────────────────────────────────────
    \multicolumn{2}{@{}l@{}}{\textbf{\textsc{Push Sand}}} \\[0.5pt]
    \rotatebox{90}{\textbf{InD}} & \includegraphics[width=\linewidth]{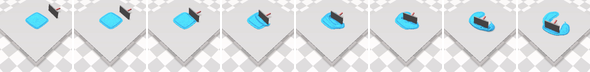} \\[0.3pt]
    \rotatebox{90}{\textbf{OoD}} & \includegraphics[width=\linewidth]{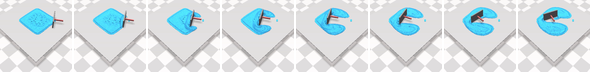} \\[3pt]
    \multicolumn{2}{@{}l@{}}{\textbf{\textsc{Pour Water}}} \\[0.5pt]
    \rotatebox{90}{\textbf{InD}} & \includegraphics[width=\linewidth]{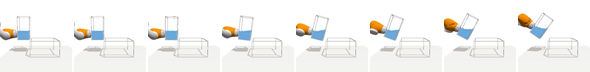} \\[0.3pt]
    \rotatebox{90}{\textbf{OoD}} & \includegraphics[width=\linewidth]{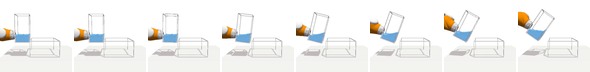} \\[3pt]
    %% Kinematics ──────────────────────────────────────────────────────────
    \multicolumn{2}{@{}l@{}}{\textbf{\textsc{Robot Arm}}} \\[0.5pt]
    \rotatebox{90}{\textbf{InD}} & \includegraphics[width=\linewidth]{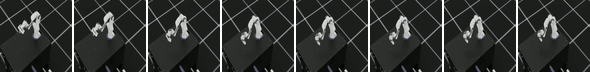} \\[0.3pt]
    \rotatebox{90}{\textbf{OoD}} & \includegraphics[width=\linewidth]{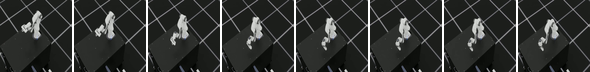} \\[3pt]
    \multicolumn{2}{@{}l@{}}{\textbf{\textsc{Reacher}}} \\[0.5pt]
    \rotatebox{90}{\textbf{InD}} & \includegraphics[width=\linewidth]{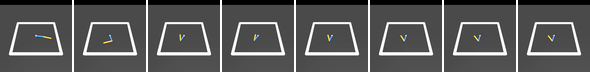} \\[0.3pt]
    \rotatebox{90}{\textbf{OoD}} & \includegraphics[width=\linewidth]{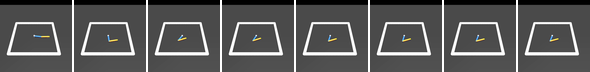} \\
  \end{tabular}
\end{minipage}
\end{center}}

  \caption{%
    \textbf{Dataset visualizations for all eight ACWM-Phys environments.}
    Left: rigid-body and deformable tasks. Right: particle and kinematics tasks.
    For each environment, InD (top) and OoD (bottom) ground-truth frames
    are shown at eight evenly-spaced timesteps from a representative episode.
  }
  \label{fig:dataset_viz}
\end{figure}

Figure~\ref{fig:dataset_viz} shows representative ground-truth frames from
all eight ACWM-Phys environments for both in-distribution (InD) and
out-of-distribution (OoD) test splits.
Each row displays eight evenly-spaced frames from a single episode,
illustrating the diversity of physical interactions and the nature of
the OoD distribution shift in each environment.

%% ─────────────────────────────────────────────────────────────────────────────
\subsection{Per-Environment Case Studies}
\label{app:case_studies}

For each environment, we show one in-distribution (InD) and one out-of-distribution (OoD)
episode with GT (top) and Pred (bottom) rows at four evenly-spaced timesteps.
For kinematics environments (Robot Arm, Reacher), an additional Overlay row blends the
GT at 45\% opacity with a blue tint over the prediction, making positional errors
directly visible without relying on side-by-side comparison alone.

\paragraph{Push Rope.}
The model faithfully predicts rope deformation under InD conditions.
Under OoD stiffness shifts, the predicted rope shape diverges from ground truth at
later timesteps, with the model underestimating rope rigidity and producing
slightly straighter configurations than observed.

\begin{figure}[h]
  \centering
  {\setlength{\tabcolsep}{2pt}\renewcommand{\arraystretch}{0.8}
\begin{center}
\begin{tabular}{r cc}
  & \textbf{In-Distribution} & \textbf{Out-of-Distribution} \\[2pt]
  \rotatebox{90}{\scriptsize GT} &
  \includegraphics[width=0.46\linewidth]{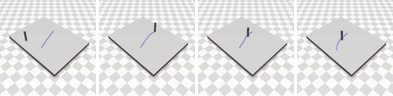} &
  \includegraphics[width=0.46\linewidth]{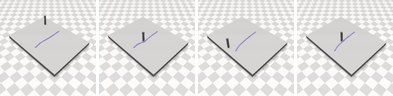} \\[1pt]
  \rotatebox{90}{\scriptsize Pred} &
  \includegraphics[width=0.46\linewidth]{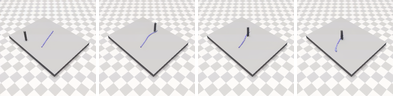} &
  \includegraphics[width=0.46\linewidth]{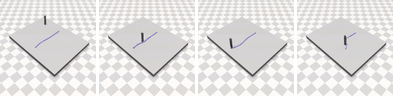} \\
\end{tabular}
\end{center}}

  \caption{\textbf{Push Rope case study.} InD (left) and OoD with longer rope (right).}
  \label{fig:cs_push_rope}
\end{figure}

\paragraph{Cloth Move.}
The model does not capture the dynamics well for both InD and OoD.
Under OoD cloth sizes (smaller or larger than the training range), the model
hallucinates incorrect cloth extents and misplaces the deformation boundary at
the sphere, producing plausible but geometrically inaccurate draping.

\begin{figure}[h]
  \centering
  {\setlength{\tabcolsep}{2pt}\renewcommand{\arraystretch}{0.8}
\begin{center}
\begin{tabular}{r cc}
  & \textbf{In-Distribution} & \textbf{Out-of-Distribution} \\[2pt]
  \rotatebox{90}{\scriptsize GT} &
  \includegraphics[width=0.46\linewidth]{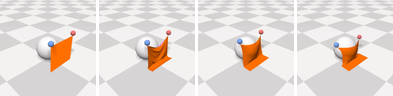} &
  \includegraphics[width=0.46\linewidth]{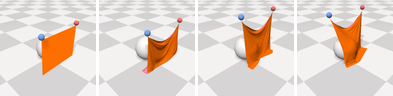} \\[1pt]
  \rotatebox{90}{\scriptsize Pred} &
  \includegraphics[width=0.46\linewidth]{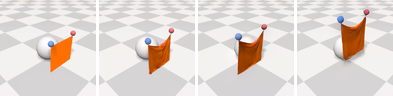} &
  \includegraphics[width=0.46\linewidth]{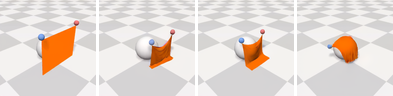} \\
\end{tabular}
\end{center}}

  \caption{\textbf{Cloth Move case study.} InD (left) and OoD cloth-size shift (right).}
  \label{fig:cs_cloth_move}
\end{figure}

\paragraph{Push Sand.}
The model generalizes partially to OoD doubled particle counts: overall granular flow
direction and pile topology are qualitatively preserved.
However, the model sometimes predicts fewer particles than are present---the sand
pile appears to \emph{shrink}---reflecting that precise particle count is not
internalized and degrades under large distribution shifts in granular density.

\begin{figure}[h]
  \centering
  {\setlength{\tabcolsep}{2pt}\renewcommand{\arraystretch}{0.8}
\begin{center}
\begin{tabular}{r cc}
  & \textbf{In-Distribution} & \textbf{Out-of-Distribution} \\[2pt]
  \rotatebox{90}{\scriptsize GT} &
  \includegraphics[width=0.46\linewidth]{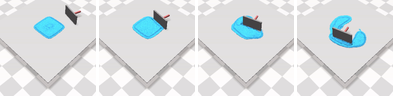} &
  \includegraphics[width=0.46\linewidth]{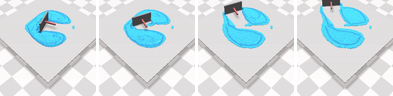} \\[1pt]
  \rotatebox{90}{\scriptsize Pred} &
  \includegraphics[width=0.46\linewidth]{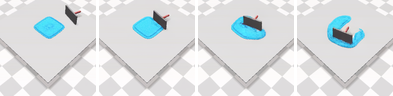} &
  \includegraphics[width=0.46\linewidth]{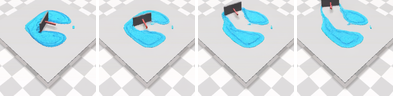} \\
\end{tabular}
\end{center}}

  \caption{\textbf{Push Sand case study.} InD (left) and OoD doubled-particle-count (right).}
  \label{fig:cs_push_sand}
\end{figure}

\paragraph{Stack Cube.}
InD stacking trajectories e.g. pick-up, transport, and placement are accurately
predicted. Under OoD target placement shifts, the model predicts a plausible
but positionally incorrect stack, indicating limited spatial extrapolation
beyond training placement configurations.

\begin{figure}[h]
  \centering
  {\setlength{\tabcolsep}{2pt}\renewcommand{\arraystretch}{0.8}
\begin{center}
\begin{tabular}{r cc}
  & \textbf{In-Distribution} & \textbf{Out-of-Distribution} \\[2pt]
  \rotatebox{90}{\scriptsize GT} &
  \includegraphics[width=0.46\linewidth]{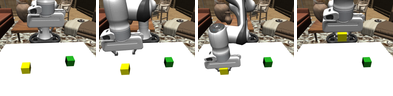} &
  \includegraphics[width=0.46\linewidth]{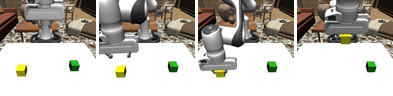} \\[1pt]
  \rotatebox{90}{\scriptsize Pred} &
  \includegraphics[width=0.46\linewidth]{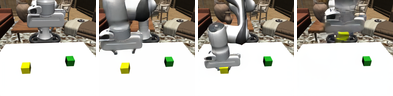} &
  \includegraphics[width=0.46\linewidth]{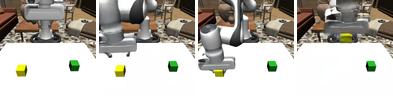} \\
\end{tabular}
\end{center}}

  \caption{\textbf{Stack Cube case study.} InD (left) and OoD placement-shift (right).}
  \label{fig:cs_stack_cube}
\end{figure}

\paragraph{Robot Arm.}
The overlay row (blue-tinted GT ghost over prediction) reveals systematic
end-effector position errors under OoD workspace expansion.
InD predictions closely match GT joint-angle trajectories; OoD predictions
reproduce plausible arm motion but with a consistent spatial offset, consistent
with the large $\Delta$PSNR and $\Delta$SSIM reported in Table~\ref{tab:eval_main}.

\begin{figure}[h]
  \centering
  {\setlength{\tabcolsep}{2pt}\renewcommand{\arraystretch}{0.8}
\begin{center}
\begin{tabular}{r cc}
  & \textbf{In-Distribution} & \textbf{Out-of-Distribution} \\[2pt]
  \rotatebox{90}{\scriptsize GT} &
  \includegraphics[width=0.46\linewidth]{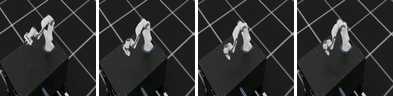} &
  \includegraphics[width=0.46\linewidth]{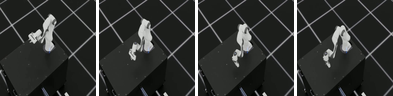} \\[1pt]
  \rotatebox{90}{\scriptsize Pred} &
  \includegraphics[width=0.46\linewidth]{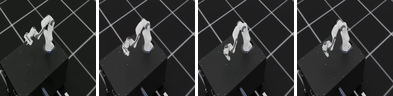} &
  \includegraphics[width=0.46\linewidth]{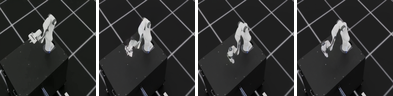} \\[1pt]
  \rotatebox{90}{\scriptsize\color{blue!60!black}Overlay} &
  \includegraphics[width=0.46\linewidth]{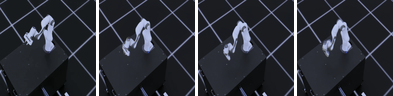} &
  \includegraphics[width=0.46\linewidth]{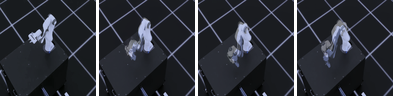} \\
\end{tabular}
\end{center}}

  \caption{%
    \textbf{Robot Arm case study.} InD (left) and OoD workspace-expansion (right).
    Overlay row: GT (blue tint, 45\% opacity) over prediction highlights positional error.
  }
  \label{fig:cs_robot_arm}
\end{figure}

\paragraph{Reacher.}
The model achieves near-perfect prediction for the two-link planar arm both InD and OoD.
The overlay row confirms negligible positional error even under OoD corner-sector
goals unseen during training, consistent with the minimal $\Delta$SSIM\,$=$\,0.000
in the main evaluation.

\begin{figure}[h]
  \centering
  {\setlength{\tabcolsep}{2pt}\renewcommand{\arraystretch}{0.8}
\begin{center}
\begin{tabular}{r cc}
  & \textbf{In-Distribution} & \textbf{Out-of-Distribution} \\[2pt]
  \rotatebox{90}{\scriptsize GT} &
  \includegraphics[width=0.46\linewidth]{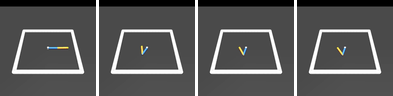} &
  \includegraphics[width=0.46\linewidth]{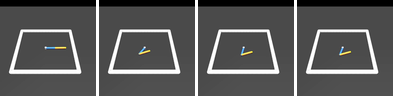} \\[1pt]
  \rotatebox{90}{\scriptsize Pred} &
  \includegraphics[width=0.46\linewidth]{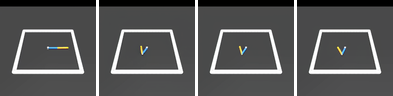} &
  \includegraphics[width=0.46\linewidth]{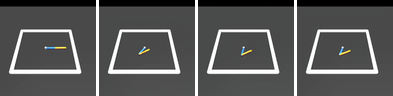} \\[1pt]
  \rotatebox{90}{\scriptsize\color{blue!60!black}Overlay} &
  \includegraphics[width=0.46\linewidth]{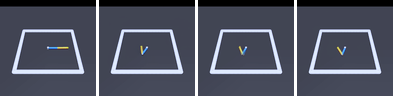} &
  \includegraphics[width=0.46\linewidth]{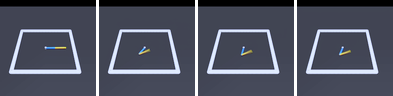} \\
\end{tabular}
\end{center}}

  \caption{%
    \textbf{Reacher case study.} InD (left) and OoD corner-sector goals (right).
    Overlay nearly coincides with Pred, confirming strong geometric generalization.
  }
  \label{fig:cs_reacher}
\end{figure}

\end{document}